\documentclass[twoside,11pt]{article}

\usepackage{xcolor}

\definecolor{linkblue}{RGB}{0,90,180}
\definecolor{lightlinkblue}{RGB}{70,150,230} 

\usepackage[
    pdftex,
    colorlinks=true,
    bookmarksopen=true,
    linkcolor=lightlinkblue,
    citecolor=linkblue,      
    urlcolor=linkblue        
]{hyperref}
\usepackage{xspace}
\usepackage[
    a4paper,
    left=3cm,
    right=3cm,
    top=3cm,
    bottom=3cm
]{geometry}

\usepackage{booktabs}
\usepackage[T1]{fontenc}
\usepackage[utf8]{inputenc}
\usepackage{lmodern}
\usepackage{libertine}
\usepackage{makecell}
\usepackage{bbm}
\usepackage{enumitem}
\usepackage{mathrsfs}
\usepackage{amssymb}
\usepackage{amsmath}
\usepackage{upgreek}
\usepackage{mathtools}
\usepackage{multirow,multicol}
\usepackage{amsthm}
\usepackage{comment}
\usepackage[font=small,labelfont=bf]{caption}
\usepackage{cleveref}
\theoremstyle{plain}
\newtheorem{theorem}{Theorem}[section]

\newtheorem{proposition}[theorem]{Proposition}
\newtheorem{lemma}[theorem]{Lemma}
\newtheorem{corollary}[theorem]{Corollary}
\newtheorem{assumption}{Assumption}

\newtheorem{remark}[theorem]{Remark}
\newtheorem{example}{Example}

\usepackage{graphicx} 
\usepackage{quoting}
\usepackage{subcaption}
\usepackage[numbers,sort&compress]{natbib}

\usepackage{lineno}

\def\R{\mathbb R}
\def\E{\mathbb E}

\def\to{\rightarrow}
\newcommand{\diff}{\mathrm{d}}

\DeclareMathOperator*{\argmin}{argmin}
\DeclareMathOperator*{\esssup}{ess\,sup}

\newcommand{\bff}{\mathbf{f}}
\newcommand{\bfF}{\mathbf{F}}

\newcommand{\bfx}{\mathbf{x}}

\newcommand{\bfy}{\mathbf{y}}

\newcommand{\bfp}{\mathbf{p}}

\usepackage{pifont}

\title{Convergence, design and training of continuous-time dropout as a random batch method}

\author{
Antonio Álvarez-López\textsuperscript{1} \qquad
Martín Hernández\textsuperscript{2}
}

\date{\vspace{-1cm}}

\begin{document}

\maketitle

\footnotetext[1]{
Departamento de Matemáticas, Universidad Autónoma de Madrid,
28049 Madrid, Spain.
}

\footnotetext[2]{
Department of Statistics and Data Science, University of California, Los Angeles, 
CA 90095 Los Angeles, USA.
}

\begingroup
\renewcommand\thefootnote{}
\footnotetext{
Source code at 
\url{https://github.com/antonioalvarezl/2026-CTDropout-RBM/}.
}
\addtocounter{footnote}{-1}
\endgroup

\renewcommand{\thefootnote}{\fnsymbol{footnote}}
\renewcommand{\thefootnote}{\arabic{footnote}}

\begin{abstract}
We study continuous-time dropout in controlled differential equations. We introduce a random-batch approximation of additive vector fields. On each time interval of length $h$, a random subset of components is activated and rescaled by its inclusion probabilities, yielding an unbiased approximation of the full field. We prove a uniform-in-time mean-square trajectory error of order $\mathcal O(h)$. At the distribution level, we derive Wasserstein and pointwise density estimates, together with global $L^1$ bounds under moment assumptions. For supervised training, we prove uniform $\mathcal O(\sqrt h)$ root-mean-square fluctuations of the randomized objective and an $\mathcal O(h)$ weak error for its expectation, leading to consistency of optimal values and near-minimizers and, under quadratic growth, convergence in control space. We further characterize how the sampling law affects the error constant through inclusion probabilities and field variance, compare fixed-size sampling, fixed partitions, and Bernoulli dropout, and derive a work--accuracy model for selecting the switching scale. Numerical experiments with neural ODEs illustrate trajectory and transport errors, sampling effects, and objective consistency at a fixed control.

\vspace{0.3cm}
\end{abstract}

\setcounter{tocdepth}{2}
\tableofcontents
\vspace{0.5cm}

\section{Introduction}

Randomized approximations offer a natural approach to reducing the computational cost of differential systems with additively decomposed vector fields. Such additive structures arise naturally in neural ODEs and related
continuous-time learning models \cite{ChenRBD18,weinan,haber}, where
evaluating the full vector field can become increasingly expensive
as the number of units grows. This raises a  numerical question: can one evolve the system using only a random subset of its components over short time intervals while retaining quantitative control of the error?

In neural networks, random masking is most commonly associated with dropout, originally introduced as a regularization mechanism \cite{srivastava2014dropout,pmlrwan13}. Our perspective is different but complementary. We study a continuous-time masking mechanism primarily as a randomized approximation of a prescribed deterministic dynamics rather than as a regularizer. In the neural ODE setting, this amounts to activating a subset of the component fields at a given time without modifying the underlying full model. Thus, the core mechanism behind dropout can also be interpreted as a randomized, reduced-cost approximation of the full dynamics.

From a numerical analysis perspective, this construction is closely related to random-batch methods and, more broadly, randomized splitting strategies. Our contribution is to formulate and analyze this type of randomized approximation for controlled vector fields, exploring its implications not only for trajectories but also for transported distributions, training objectives, and sampling design.

Related stochastic interpretations of residual and continuous-time neural models have been investigated through noise-injected dynamics and neural SDEs \cite{Sun2018StochasticTO,Liu2020_Neural_SDE}, while more recent work considers the direct random activation and deactivation of neurons \cite{lee2026neural}. Dropout has also been studied from the perspective of training dynamics; in particular, recent mean-field analyses identify distinct asymptotic regimes depending on the joint scaling of width, learning rate, and dropout rate \cite{chizat2025phasediagramdropouttwolayer}. Our focus, however, is on approximation: starting from a given controlled dynamics, how accurately can it be replaced by a random-batch counterpart? How does the resulting error propagate to transport and training objectives? And how should the randomization be designed to optimally balance accuracy and computational work?

\subsection{Contributions}

Our main contributions are the following:

\begin{enumerate}

\item \textbf{Random-batch formulation and trajectory convergence.}
We introduce a general random-batch construction for controlled additive vector fields that includes Bernoulli dropout as a special case. Given a switching scale $h$, we prove a uniform-in-time mean-square trajectory error of order $\mathcal{O}(h)$. We also obtain a stability estimate for cases where the full and randomized systems are driven by different controls.

\item \textbf{Transport of distributions.}
We extend our trajectory analysis to probability distributions transported by the corresponding flows. We establish an $\mathcal{O}(\sqrt{h})$ Wasserstein estimate, an order-$h$ pointwise mean-square estimate for the transported densities, and global $L^1$ bounds under moment assumptions. For compactly supported initial densities, the $L^1$ convergence rate is $\mathcal{O}(\sqrt{h})$.

\item \textbf{Consistency of training objectives.}
Uniformly over controls taking values in a compact set, we establish an $\mathcal O(\sqrt h)$ root-mean-square error for the single-schedule objective and an $\mathcal O(h)$ weak error for its expectation. These estimates imply consistency of optimal values and near-minimizers. Under quadratic growth and an optimization tolerance of order $\mathcal O(h)$, the distance to the full minimizer set is $\mathcal O(\sqrt h)$ in control space. If the minimizer is unique, this gives convergence to that control; an additional parameter-Lipschitz assumption yields $\mathcal O(h)$ mean-square convergence of the corresponding randomized trajectories.

\item \textbf{Sampling design and work--accuracy trade-off.}
Although the mean-square trajectory error scales as $\mathcal{O}(h)$ for any admissible sampling scheme, its prefactor is highly dependent on the chosen sampling law. We identify the field variance and the minimal inclusion probability as the critical parameters and compare uniform fixed-size sampling, fixed partitions, and Bernoulli dropout. We then derive a work--accuracy model that determines the optimal switching scale for a prescribed tolerance and characterizes the regimes in which random batching is computationally preferable within this work model.

\end{enumerate}

\subsection{Related work}

\paragraph{\textbf{Random batch methods and randomized splitting.}}
Splitting methods approximate composite dynamics by decomposing them into simpler subproblems, with classical examples including Lie--Trotter-type schemes \cite{trotter59,MacNamara2016,LANSER1999201}. Random batch methods instead use stochastic subsampling to reduce the cost of evaluating large systems. Originally developed for interacting-particle systems \cite{JIN2020108877,MR4230431}, they have since been applied to optimal control and model predictive control \cite{MR4433122,veldman2023stability}, as well as PDEs and related settings \cite{MR4361973,MR4436794,MR4744252,eisenmann2022randomized,hernandezcorella2025,discretizedRBM,continuousRBM}. The present paper adapts this random-batch viewpoint to additive controlled vector fields, framing it as continuous-time dropout and leading to additional questions concerning transported distributions, training objectives, and sampling design. 

\paragraph{\textbf{Dropout and stochastic neural dynamics.}}
Dropout was introduced as a regularization mechanism for neural networks \cite{srivastava2014dropout,pmlrwan13} and has subsequently been studied from both Bayesian and optimization perspectives \cite{gal2017concretedropout,NIPS2015_bc731692,mianjy2020convergence}. For residual and continuous-time architectures, stochastic perturbations have been linked to stochastic differential equations (SDEs) and continuous-time models \cite{Sun2018StochasticTO,Liu2020_Neural_SDE}. More recently, continuous-time neuron activation and deactivation has been modeled directly through random switching \cite{lee2026neural}. Furthermore, recent mean-field studies of training dynamics under dropout in wide two-layer networks have identified distinct asymptotic regimes \cite{chizat2025phasediagramdropouttwolayer}. In contrast, our analysis treats dropout-type masking primarily as a numerical approximation of a prescribed deterministic controlled system.

\paragraph{\textbf{Pruning and sparsification.}}
Pruning provides a complementary mechanism for reducing neural-network complexity; classical approaches include \cite{LeCun1990,Hassibi1993,Han2015,Han2016}, and we refer to \cite{ChengSurveyPruning2024} for a broader overview. Sparsification has also been explored in continuous-depth models \cite{liebenwein2021sparse,pmlr-v202-mo23c}. Unlike pruning, the random-batch construction considered here does not permanently discard components: the full model is preserved, while different subsets are dynamically activated along the flow. In contrast to these approaches, our framework also provides convergence guarantees.

\subsection{Roadmap}

The paper is organized as follows.
\Cref{sec:math_formulation} introduces the controlled dynamics and their random-batch approximation.
\Cref{sec:mainresults} establishes uniform-in-time trajectory estimates, their consequences for measure transport, and strong and weak consistency results for the random-batch training objectives.
\Cref{sec:design} compares canonical sampling schemes through their inclusion probabilities and field variance (\Cref{ss:varsamp}), and develops a work--accuracy model for selecting the switching scale (\Cref{sec:cost_accuracy}).
\Cref{sec:numerical_results} examines trajectory errors, sampling effects, objective consistency and Monte Carlo averaging at a fixed control, measure transport, and the work--accuracy trade-off.
\Cref{sec:conclusions} summarizes the results and discusses open directions.
All proofs are collected in \Cref{sec:proof_of_the_results}. Numerical implementation details are collected in \Cref{sec:numerical_details}.

\subsection{Notation}\label{ss:notation}
Let $[n]\coloneqq \{1,\dots,n\}$. Scalars, vectors, and matrices are denoted by plain, bold, and uppercase letters, respectively. For $x,y\in\R$, $x\wedge y\coloneqq \min\{x,y\}$ and $x\vee y \coloneqq \max\{x,y\}$. Let $B_R\subset\R^d$ be the closed origin-centered ball of radius $R$. The Euclidean inner product is $\langle \cdot,\cdot\rangle$, with induced vector norm $\|\cdot\|$. For matrices, $\|\cdot\|$ denotes the operator norm. 

Regarding functions, subscripts indicate time: $f_t(\cdot)\coloneqq f(t,\cdot)$. We use standard notation for Lebesgue spaces $L^p$ ($1\le p\le\infty$) and absolutely continuous curves ${AC}([0,T];\R^d) 
$. We denote by ${C}^{0,1}$ the space of globally Lipschitz functions (with constant $\lambda_{\bff}$), and write ${C}^{1,1}$ (resp. ${C}^{1,1}_{\rm loc}$) when $\bff\in{C}^{1}$ and $\nabla \bff$ is globally (resp. locally) Lipschitz. For a bivariate function $\bff(\bfx,\theta)$, $\lambda_{\bff,\bfx}$ and $\lambda_{\bff,\theta}$ denote its Lipschitz constants in one variable holding uniformly over the other. A third subscript restricts the domain
of uniformity (e.g., $\lambda_{\bff,\bfx,A}$ is the Lipschitz constant in
$\bfx$, uniformly for all $\theta\in A$).

\section{Formulation}\label{sec:math_formulation}
\subsection{Full model}\label{ss:nodes}

Fix $d\ge1$ and $T>0$, and consider the controlled differential equation
\begin{align}\tag{$\mathsf{FM}$}\label{eq:FM}
\begin{cases}
\dot\bfx_t&=\bfF(\bfx_t,\vartheta_t), \hspace{1cm}\text{for a.e. }t\in(0,T),\\
\bfx_0&\in\R^d,
\end{cases}
\end{align}
where the control is $\vartheta\in L^\infty(0,T;\Theta)$ with $\Theta\subseteq\R^m$. For some $p\geq1$, we assume $\Theta$ and $\bfF$ factorize as
\begin{equation}\label{eq: Fsum}
\theta=(\theta_i)_{i\in[p]}\in\prod_{i\in[p]}\Theta_i\eqqcolon\Theta,\hspace{1cm}\|\theta\|^2\coloneqq\sum_{i\in[p]}\|\theta_i\|^2, \hspace{1cm}\bfF(\bfx,\theta)=\sum_{i\in[p]}\bff_i(\bfx,\theta_i)
\end{equation}
where each $\Theta_i$ is closed, and hence so is $\Theta$.


Motivated by neural ODEs, where $\mathbf{F}$ is realized as a neural network and $p$ denotes its (constant) width, we refer to each $\bff_i:\R^d\times \Theta_i\to\R^d$ as a \emph{neuron}. For instance, when $\bfF$ is a single-layer neural network with time-dependent parameters, \eqref{eq:FM} takes the form
\begin{equation}\tag{$\mathsf{NODE}$}\label{eq:node-p}
\begin{cases}
\dot\bfx_t&=\sum_{i\in[p]} \mathbf{w}_{i,t}\sigma(\langle\mathbf{a}_{i,t},\bfx_t\rangle+b_{i,t}),\hspace{1cm} \text{for a.e. }t\in (0,T),\\
\bfx_0&\in \R^d.
\end{cases}
\end{equation}
Here, $\bff_i(\bfx,\theta_i) \coloneqq \mathbf w_i \sigma\left(\langle\mathbf a_i,\bfx\rangle+b_i\right)$ with $\sigma\in{C}^{0,1}(\R)$ and $\theta_i=(\mathbf w_i,\mathbf a_i,b_i)\in\R^d\times\R^d\times\R$, so that $\Theta=(\R^{2d+1})^p$. 
Although we focus on \eqref{eq:node-p} for the numerical experiments, the formulation of \eqref{eq:FM} also covers other deep learning models.

\subsubsection*{Well-posedness}

Throughout, we impose the following baseline regularity. For every $i\in[p]$, the field $\bff_i$ is continuous on $\R^d\times\Theta_i$, and $\bff_i$ is globally Lipschitz in $\bfx$, locally uniformly with respect to $\theta_i$. More precisely, for every compact set $\mathsf K_i\subset\Theta_i$, there exists $\lambda_{\bff_i,\bfx,\mathsf K_i}<\infty$ such that
\begin{equation}\label{ass:baseline_lipx}
\|\bff_i(\bfx,\theta_i)-\bff_i(\bfy,\theta_i)\| \le \lambda_{\bff_i,\bfx,\mathsf K_i}\|\bfx-\bfy\|\qquad \text{for all } \bfx,\bfy\in\R^d\text{ and }\theta_i\in\mathsf K_i.
\end{equation}
Therefore, since all $\Theta_i$ are closed, for each $\vartheta\in L^\infty(0,T;\Theta)$ we can define a finite aggregated Lipschitz constant
\begin{equation}\label{eq:condlipf_aggregated}
\lambda_{\bfF,\bfx,\vartheta} \coloneqq \sum_{i\in[p]} \esssup_{t\in(0,T)} \lambda_{\bff_i(\cdot,\vartheta_{i,t})}
\end{equation}
which clearly satisfies
\begin{equation}\label{eq:condlipf}
    \esssup_{t\in(0,T)}\|\bfF(\bfx,\vartheta_t)-\bfF(\bfy,\vartheta_t)\| \leq \lambda_{\bfF,\bfx,\vartheta}\|\bfx-\bfy\| \qquad \text{for all } \bfx,\bfy\in\R^d,
\end{equation}
and prevents arbitrary cancellations among individual neurons. Moreover, since $\bff_i(0,\cdot)$ is continuous, the quantity $\kappa_{\bfF,0,\vartheta}\coloneqq \sum_{i\in[p]}\|\bff_i(0,\vartheta_{i,\cdot})\|_{L^\infty(0,T)}$ is finite. This yields the uniform linear growth bound:
\begin{equation}\label{eq:condgrowthf}
    \esssup_{t\in(0,T)}\|\bfF(\bfx,\vartheta_t)\| \leq \lambda_{\bfF,\bfx,\vartheta}\|\bfx\| + \kappa_{\bfF,0,\vartheta} \qquad \text{for all } \bfx\in\R^d.
\end{equation}
These conditions satisfy all the hypotheses of \Cref{lemma:caratheodory}. Consequently, for any initial state $\bfx_0\in\R^d$ and control $\vartheta\in L^\infty(0,T;\Theta)$, there exists a unique absolutely continuous solution $\bfx(\cdot)\in {AC}([0,T];\R^d)$ to the forward model \eqref{eq:FM}.

\subsubsection*{Standing assumptions}
Additional assumptions on the functions $\bff_i$ will be invoked only where used. For the reader's convenience and to fix notation, we collect them here:

\begin{assumption}\label{ass:menu}
For each $i\in[p]$, we consider the following Lipschitz conditions (recall the notation introduced in \Cref{ss:notation}):
\begin{align}
0\le\lambda_{\bff_i,\theta,\mathsf B,\mathsf K_i}<\infty
&\qquad
\text{for all compact sets $\mathsf B\subset\R^d$ and
$\mathsf K_i\subset\Theta_i$,}
\tag{A1}\label{ass:lipthetaF}\\
0\le\lambda_{\nabla_\bfx\bff_i,\bfx,\mathsf K_i}<\infty
&\qquad
\text{for every compact set $\mathsf K_i\subset\Theta_i$.}
\tag{A2}\label{ass:lipxdxF}
\end{align}
That is, $\bff_i$ is locally Lipschitz in $\theta$ (locally uniformly in $\bfx$) \emph{(A1)}, while $\nabla_\bfx\bff_i$ is globally Lipschitz in $\bfx$ (locally uniformly in $\theta$) \emph{(A2)}. Each result will state which of these assumptions it requires.
\end{assumption}

Whenever these properties hold for every $\bff_i$,
the full field $\bfF$ inherits the corresponding property. In particular,
if $\mathsf K\subset\Theta$ is compact with
$\mathsf K_i\coloneqq\operatorname{proj}_i\mathsf K$, we explicitly define the
corresponding aggregated constants as in \eqref{eq:condlipf_aggregated}:
\[
\lambda_{\bfF,\theta,\mathsf B,\mathsf K} \coloneqq \sum_{i\in[p]}\lambda_{\bff_i,\theta,\mathsf B,\mathsf K_i},
\qquad
\lambda_{\nabla_\bfx\bfF,\bfx,\mathsf K} \coloneqq \sum_{i\in[p]}\lambda_{\nabla_\bfx\bff_i,\bfx,\mathsf K_i}.
\]

\begin{example}
For \eqref{eq:node-p}, we have $\|\bff_i(\bfx,\theta_i)-\bff_i(\bfy,\theta_i)\| \leq \lambda_\sigma \|\mathbf{w}_i\|\|\mathbf{a}_i\|\|\bfx-\bfy\|$, which ensures the baseline spatial regularity for any compact $\mathsf K_i$. \Cref{ass:lipthetaF} follows from $\sigma$ being Lipschitz, while \Cref{ass:lipxdxF} requires a smoother activation like GELU ($\sigma\in{C}^{1,1}(\R)$), rather than ReLU.
\end{example}

\subsection{Random model}\label{ss:rbm}

\begin{figure}
\centering
\includegraphics[scale=0.55]{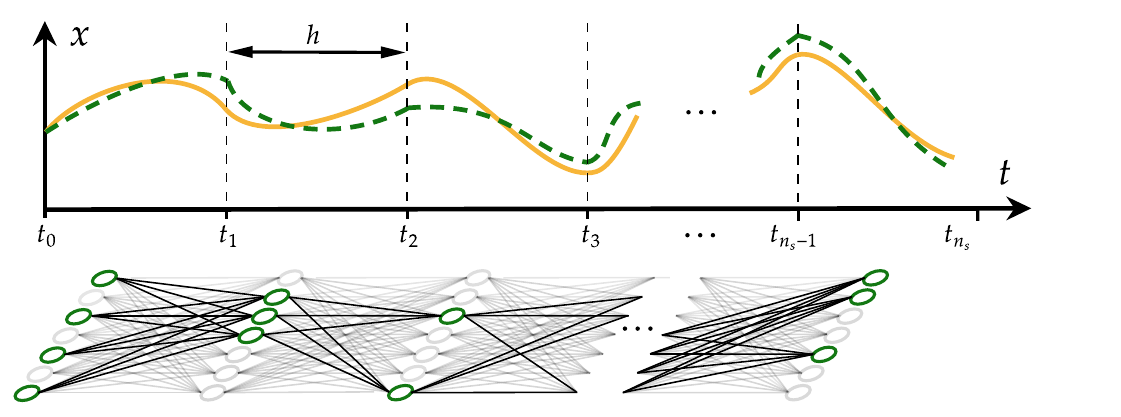}
\caption{Top: trajectories generated by the full model ({\color{orange}orange}) and the random model ({\color{green!50!black}green}). Bottom: vector fields $\mathbf{F}^{(w_{k_t})}$ obtained by retaining, on each interval $[t_{k-1},t_k)$, the neurons selected by $\omega_{k_t}$ and masking the remaining ones.}
\label{fig:rbm_node}
\end{figure}
We now introduce the standard random batch framework. Fix a number of batches $n_b\ge1$ and a number of time steps $n_s\ge1$. Consider a covering 
\begin{equation*}
\mathcal{B}_1,\dots,\mathcal{B}_{n_b}\subseteq[p],\hspace{1cm}\text{with}\hspace{1cm}  \bigcup_{j\in[n_b]} \mathcal{B}_j = [p],
\end{equation*}
and let $\omega$ be a random variable on $[n_b]$ with probabilities
\begin{equation}\label{def:qj}
q_j\coloneqq\mathbb P(\omega=j)>0, \qquad j\in[n_b], \qquad \sum_{j\in[n_b]}q_j=1.
\end{equation}
For each neuron $i\in[p]$, define its inclusion probability by
\begin{equation}\label{eq:definition_pi}
\pi_i \coloneqq \mathbb P(i\in\mathcal B_\omega) = \sum_{\ell\in[n_b]}q_\ell\mathbf{1}_{\{i\in\mathcal B_\ell\}}.
\end{equation}
Since the batches cover $[p]$ and $q_j>0$, we have
\begin{equation*}
\pi_{\min}\coloneqq\min_{i\in[p]}\pi_i>0.
\end{equation*}
For each $j\in[n_b]$, define
\begin{equation}\label{eq:Fomegaj}
\bfF^{(j)}(\bfx,\theta) \coloneqq \sum_{i\in\mathcal B_j}\frac{1}{\pi_i}\bff_i(\bfx,\theta_i).
\end{equation}
The normalization by $\pi_i$ makes the random vector field unbiased:
\begin{equation}\label{eq:unbiasedF}
\mathbb E\bigl[\bfF^{(\omega)}(\bfx,\theta)\bigr] =  \sum_{j\in [n_b]} \bfF^{(j)}(\bfx,\theta)\,q_j=\sum_{i\in[p]}\frac{\bff_i(\bfx,\theta_{i})}{\pi_i}\sum_{j\in[n_b]} q_j\mathbf{1}_{\{i \in \mathcal{B}_{j}\}}=\bfF(\bfx,\theta).
\end{equation}

Partition $[0,T]$ using $t_k=kh$ for $k=0,\dots,n_s$, with $h\coloneqq T/n_s$. Let $\omega_1,\dots,\omega_{n_s}$ be i.i.d.\ copies of $\omega$ and define $k_t\coloneqq 1+\lfloor t/h\rfloor$ for $t\in[0,T)$ and $k_T\coloneqq n_s$. For the same initial condition $\bfx_0$ and control $\vartheta$ as in \eqref{eq:FM}, define the random model
\begin{align}\tag{$\mathsf{RM}$}\label{eq:RM}
\begin{cases}
\dot{\hat{\bfx}}_t =\hat{\bfF}_t(\hat{\bfx}_t,\vartheta_t), \hspace{1cm}&\text{for a.e. }t\in(0,T),\\
\hat{\bfx}_0=\bfx_0,
\end{cases}
\end{align}
where
\begin{equation}\label{eq:hatFkt}
\hat{\bfF}_t(\bfx,\theta) \coloneqq \bfF^{(\omega_{k_t})}(\bfx,\theta) = \sum_{i\in\mathcal B_{\omega_{k_t}}} \frac{1}{\pi_i}\bff_i(\bfx,\theta_i).
\end{equation}
By \eqref{eq:unbiasedF}, $\hat{\bfF}_t$ is unbiased for a.e. $t$. The Lipschitz and growth bounds for $\bfF$ yield the corresponding bounds for $\hat{\bfF}$ with constants multiplied by $\pi_{\min}^{-1}$. For example,
\begin{equation*}
\begin{aligned}
\|\hat{\bfF}_t(\bfx,\vartheta_t) -\hat{\bfF}_t(\bfy,\vartheta_t)\| 
\le \sum_{i\in\mathcal B_{\omega_{k_t}}} \frac{1}{\pi_i} \|\bff_i(\bfx,\vartheta_{i,t}) -\bff_i(\bfy,\vartheta_{i,t})\| \le \frac{\lambda_{\bfF,\bfx,\vartheta}}{\pi_{\min}} \|\bfx-\bfy\|
\end{aligned}
\end{equation*}
holds for almost every $t\in(0,T)$. The same argument applies to \Cref{ass:lipthetaF,ass:lipxdxF}. Thus, for every realization of $(\omega_k)_{k\in[n_s]}$, \Cref{lemma:caratheodory} gives a unique solution $\hat{\bfx}\in{AC}([0,T];\R^d)$ to \eqref{eq:RM}.

\subsubsection*{Interpretation}
The batch selection in \eqref{eq:RM} is constant on each interval $[t_{k-1},t_k)$. This can be interpreted as a continuous-time analogue of inverted dropout: the neurons in $\mathcal B_{\omega_k}$ remain active during the interval, while the remaining neurons are masked.



\section{Main results}\label{sec:mainresults}

\subsection{Trajectory-level}\label{sec:convergence_dynamics}

To track variability along trajectories, for a.e. $t\in(0,T)$, define
\begin{align}\label{def:Lambda}
\Lambda_t(\bfx_0,\vartheta)\coloneqq\E_{\omega}\left[\|\bfF(\bfx_t,\vartheta_t)-\hat{\bfF}_t(\bfx_t,\vartheta_t)\|^2\right]=\sum_{j\in[n_b]}\big\|\bfF(\bfx_t,\vartheta_t)-\bfF^{(j)}(\bfx_t,\vartheta_t)\big\|^2 q_j,
\end{align}
where $\bfx_t$ solves \eqref{eq:FM} with initial data $\bfx_0$ and control $\vartheta$. As defined, $\Lambda$ is independent of $h$. Moreover, since $\bfx_t$ is continuous on $[0,T]$ and $\bfF$ satisfies the growth bound \eqref{eq:condgrowthf}, we have\footnote{Using $\|\bfF(\bfx_t,\vartheta_t)-\hat{\bfF}_t(\bfx_t,\vartheta_t)\|^2\leq 2\left(1+\pi_{\min}^{-2}\right)(\lambda_{\bfF,\bfx,\vartheta}\|\bfx_t\|+\kappa_{\bfF,0,\vartheta})^2$, derived from \eqref{eq:condgrowthf} and \eqref{eq:Fomegaj}.}
\begin{align}\label{eq: uniform_bound_Lambda}
\Lambda(\bfx_0,\vartheta)\in L^\infty(0,T),\hspace{1cm}\text{for all }\bfx_0\in\R^d\text{ and } \vartheta\in L^\infty(0,T;\Theta).
\end{align}





\medbreak 

Our first theorem shows that, for any batch decomposition, the expected squared error between trajectories decays linearly in $h$. Consequently, when $h$ is small---corresponding to rapid switching in \eqref{eq:RM}---the two systems behave similarly.

\begin{theorem}\label{th:convergence}
For any $\bfx_0\in\R^d$ and $\vartheta\in L^\infty(0,T;\Theta)$, there is a finite constant
$\mathsf S(\bfx_0,\vartheta)\ge0$, independent of $h$, such that for every $n_s \ge 1$ (with $h=T/n_s$), the solutions $\bfx$ of \eqref{eq:FM} and $\hat{\bfx}$ of \eqref{eq:RM} satisfy
\begin{equation}\label{eq:bound.th:convergence}
\max_{t\in[0,T]}\E_{\omega}\left[\|\bfx_t-\hat{\bfx}_t\|^2\right]\leq  \mathsf{S}(\bfx_0,\vartheta) \, h.  
\end{equation}
\end{theorem}
The dependence of $\mathsf S$ on the sampling design, through $\pi_{\min}$ and $\Lambda$, is made explicit in \eqref{eq:S_factorized}. $\mathsf S$ also grows at most quadratically with respect to the initial datum $\bfx_0$, as a consequence of the linear growth of $\bfF$. That is, there exists a constant $C>0$ independent of $h$ and $\bfx_0$ such that
\begin{equation}\label{eq:S_quadratic}
\mathsf{S}(\bfx_0,\vartheta) \le C\big(1+\|\bfx_0\|^2\big).
\end{equation}
Moreover, if $\mathsf K\subset\Theta$ is compact, the constants in the proof can
be chosen uniformly over $\vartheta\in L^\infty(0,T;\mathsf K)$.
Consequently, if we denote by $\bfx_{\vartheta,t}$ the state of \eqref{eq:FM} under $\vartheta$ at time $t$, and similarly by $\hat{\bfx}_{\vartheta,t}$, then for every $\bfx_0\in\R^d$ there exists $\mathsf S(\bfx_0,\mathsf K)<\infty$ such that
\[
\sup_{\vartheta\in L^\infty(0,T;\mathsf K)}
\max_{t\in[0,T]}
\E_\omega\|\bfx_{\vartheta,t}-\hat\bfx_{\vartheta,t}\|^2
\le \mathsf S(\bfx_0,\mathsf K)h.
\]


While \Cref{th:convergence} applies when the same control is used in the full and random models, the result extends to different controls under an additional Lipschitz assumption.

\begin{corollary}\label{cor:convergence_different_control}
Assume \Cref{ass:lipthetaF}. Let $\bfx_0\in\R^d$ and let
$\mathsf K\subset\Theta$ be compact. Then there exists a constant
$C_{\bfx_0,\mathsf K}>0$, independent of $h$, such that for every
$\vartheta_1,\vartheta_2\in L^\infty(0,T;\mathsf K)$ and every
$n_s\ge1$ (with $h=T/n_s$),
\[
\max_{t\in[0,T]}
\E_\omega\left[
\|\bfx_{\vartheta_1,t}-\hat\bfx_{\vartheta_2,t}\|^2
\right]
\le
C_{\bfx_0,\mathsf K}
\left(
h+
\|\vartheta_1-\vartheta_2\|_{L^1(0,T)}^2
\right).
\]
\end{corollary}

We conclude with an application of \Cref{th:convergence} to quantify the concentration of paths.

\begin{remark}
For any $\varepsilon>0$, Markov's inequality gives
\begin{equation*}
\sup_{t\in[0,T]}\mathbb{P}\left(\|\mathbf{x}_t - \hat{\mathbf{x}}_t\|^2 > \varepsilon \right)
\le \frac{1}{\varepsilon}\max_{t\in[0,T]}\E_{\omega}\left[\|\bfx_t-\hat{\bfx}_t\|^2\right] \leq \frac{\mathsf{S}(\bfx_0,\vartheta) }{\varepsilon}\,h.
\end{equation*}
Thus, for fixed parameters and batch decomposition, as the switching becomes infinitely fast, the random state $\hat{\bfx}_t$ converges to the full state $\bfx_t$ in probability at each time, with a probability bound uniform over the interval $[0,T]$.
\end{remark}

\subsection{Distribution-level}\label{ss:transp}
We have quantified how dropout perturbs the forward evolution of a single datum $\bfx_0$. We now assume that $\bfx_0$ has law $\diff\mu_{\rm B}=\rho_{\rm B}\diff\bfx$, where $\rho_{\rm B}\in{C}^{0,1}(\R^d)\cap L^1(\R^d)$ is a non-negative probability density (i.e., with unit mass). When the trajectories evolve according to \eqref{eq:FM}, their density $\rho_t$ solves the continuity equation
\begin{equation}\label{eq: neurtransp}
\begin{cases}
\partial_t\rho_t(\bfx) + \nabla_\bfx\cdot \bigl(\bfF(\bfx,\vartheta_t)\rho_t(\bfx)\bigr) = 0,\hspace{1cm} &\text{in }(0,T)\times \R^d, \\
\rho_0(\bfx)= \rho_{\rm B}(\bfx), \hspace{1cm} &\text{on }\R^d.
\end{cases}
\end{equation}
Besides \eqref{eq:condlipf}, assume that \Cref{ass:lipxdxF} holds. Then \eqref{eq: neurtransp} has a unique weak solution 
\begin{equation*}
    \rho\in {C}^0([0,T]\times\R^d) \cap L^\infty\bigl(0,T; {C}^{0,1}(\R^d)\bigr).
\end{equation*}
Indeed, \Cref{ass:lipxdxF} combined with \eqref{eq:condlipf} ensures that $\bfF(\cdot,\vartheta_t)\in{C}^{1,1}(\R^d;\R^d)$ with uniformly bounded spatial gradients. The flow $\Phi_t$ of \eqref{eq:FM} (with $\Phi_0=\operatorname{Id}$ and $\Phi_t(\bfx_0)=\bfx_t$) is therefore a global ${C}^1$-diffeomorphism on $\R^d$, and the density satisfies
\begin{equation}\label{eq:Liouville}
\rho_t(\bfx) = \rho_{\rm B}\bigl(\Phi_{t}^{-1}(\bfx)\bigr)\exp\left(-\int_0^t \nabla_\bfx\cdot\bfF\big(\Phi_s(\Phi_t^{-1}(\bfx)),\vartheta_s\big)\diff s\right).
\end{equation}
The dropout analogue of \eqref{eq: neurtransp} replaces $\bfF$ by $\hat\bfF$:
\begin{equation}\label{eq: neurtransprb}
\begin{cases}
\partial_t\hat{\rho}_t(\bfx) + \nabla_\bfx\cdot \left(\hat{\bfF}_t(\bfx,\vartheta_t)\hat{\rho}_t(\bfx)\right) = 0, \hspace{1cm}&\text{in }(0,T)\times \R^d,\\
\hat{\rho}_0 (\bfx)= \rho_{\rm B}(\bfx), &\text{on }\R^d.
\end{cases}
\end{equation}
Existence and uniqueness for \eqref{eq: neurtransprb} follow verbatim from the same argument, with an expression for $\hat{\rho}_t$ analogous to \eqref{eq:Liouville}, replacing $\Phi_t$ by the random flow $\hat\Phi_t$.

\begin{remark}\label{rem:W2}
Let $\mathcal{P}_2(\R^d)$ be the space of probability measures with finite second moments. For $\mu,\nu\in\mathcal{P}_2(\R^d)$, the $2$-Wasserstein distance is defined by
\begin{equation*}
W_2(\mu,\nu)\coloneqq\left(\inf_{\pi\in\Pi(\mu,\nu)}\int_{\R^d\times\R^d}\|\bfx-\bfy\|^2\diff\pi(\bfx,\bfy)\right)^{1/2},
\end{equation*}
where $\Pi(\mu,\nu)\coloneqq\{\pi\in\mathcal{P}(\R^d\times\R^d): \pi(\cdot,\R^d)=\mu,\ \pi(\R^d,\cdot)=\nu\}$. Assume that $\mu_{\rm B}\in\mathcal{P}_2(\R^d)$. By virtue of \Cref{th:convergence}, it holds $\E_{\omega}[\|\Phi_t(\bfx_0)-\hat\Phi_t(\bfx_0)\|^2]\le \mathsf{S}(\bfx_0,\vartheta)\, h$. Therefore, for $\mu_t\coloneqq(\Phi_t)_{\#}\mu_{\rm B}$ and $\hat\mu_t\coloneqq(\hat\Phi_t)_{\#}\mu_{\rm B}$, Jensen's inequality yields
\begin{equation*}
\E_{\omega}\left[W_2(\mu_t,\hat\mu_t)\right] \le \sqrt{h}\left(\int _{\R^d}\mathsf{S}(\bfx_0,\vartheta)\diff\mu_{\rm B}(\bfx_0)\right)^{1/2}.
\end{equation*}
\end{remark}

The next result provides a pointwise error estimate, obtained by tracing backward in time the two characteristics that end at $\bfx$.

\begin{theorem}\label{thm:transport} 
Assume \Cref{ass:lipxdxF}. For any probability density $\rho_{\rm B}\in {C}^{0,1}(\R^d)$ and $\vartheta\in L^\infty(0,T;\Theta)$, there exists a constant $C(\rho_{\rm B},\vartheta)>0$, independent of $h$, such that for every $n_s\ge1$ (with $h=T/n_s$), the solutions $\rho$ and $\hat \rho$ of \eqref{eq: neurtransp}--\eqref{eq: neurtransprb} satisfy
\begin{align}\label{eq: boundrho}
 \E_{\omega}\left[ |\rho_t(\bfx)-\hat{\rho}_t(\bfx)|^2 \right] \leq C(\rho_{\rm B},\vartheta)\big(1+\|\bfx\|^2\big)h,
\end{align}
for all $(t,\bfx)\in[0,T]\times\R^d$. 
\end{theorem}

Although the estimate in \Cref{thm:transport} is pointwise, its right-hand side grows at most quadratically in space. By imposing additional moment assumptions on the initial distribution, we can control the tail behavior and integrate this pointwise bound to obtain error estimates in global metrics. In particular, the total variation distance---which equals one half of the $L^1$-distance between the corresponding densities---can be bounded as follows:

\begin{corollary}\label{cor:L1bound}
Assume the hypotheses of \Cref{thm:transport}. If $\rho_{\rm B}$ has finite $q$-th moment for some $q>0$, then there exists $C(\rho_{\rm B},\vartheta,T,q)>0$, independent of $h$, such that for every $n_s\ge1$ (with $h=T/n_s$) and every $t\in[0,T]$,
\begin{equation}\label{eq:L1rate}
\E_{\omega}\left[\|\rho_t-\hat\rho_t\|_{L^1(\R^d)}\right] \le C(\rho_{\rm B},\vartheta,T,q) h^{\frac{q}{2q+2d+2}}.
\end{equation}
In particular, if $\rho_{\rm B}$ is compactly supported, the convergence rate becomes $\mathcal{O}(\sqrt{h})$.
\end{corollary}

\subsection{Training consistency}\label{ss:train}

Supervised learning with continuous-depth models can be formulated as an optimal control problem; see, e.g., \cite{weinanoptimalcontrol,EGePiZua}. Consider the forward system
\begin{equation}\label{eq:forward_full}
\begin{cases}
\dot{\bfx}_{m,t} = \sum_{i\in[p]} \bff_i(\bfx_{m,t},\vartheta_{i,t}), \hspace{1cm} \text{for a.e. } t\in (0,T),\\
\bfx_{m,0}=\bfx_{m},
\end{cases}
\end{equation}
for $m\in[n_d]$ and fixed inputs $\bfx_{m}\in\R^d$. We restrict the optimization to an admissible class $\mathcal U = L^\infty(0,T;\mathsf K)$, where $\mathsf K \subset \Theta$ is compact and convex. This compactness ensures that the Lipschitz and growth constants of the vector fields hold uniformly across all $\vartheta\in\mathcal U$. We denote by $\bfx_{m,t}\equiv\bfx_{m,\vartheta,t}$ the corresponding solutions. The objective functional of the full model is
\begin{equation}\label{eq:objective_full}
J(\vartheta) \coloneqq \frac{\alpha}{2}\|\vartheta\|_{L^2(0,T;\R^m)}^2 + \sum_{m\in[n_d]}g_m(\bfx_{m,T}) + \beta\sum_{m\in[n_d]} \int_0^T \ell_m(\bfx_{m,t}) \diff t,
\end{equation}
where $\alpha>0$ and $\beta\geq0$ are fixed, and $\ell_m, g_m \in {C}^{1,1}_{\rm loc}(\R^d; \R_{\geq0})$ for each $m\in[n_d]$. 

Following \Cref{ss:rbm}, we incorporate random batching to obtain the randomized trajectories $\hat\bfx_{m,t}$. For a given deterministic control $\vartheta$ and a realization of the batch sequence $\omega$, the randomized objective is
\begin{equation}\label{eq:objective_random}
\hat{J}_h(\vartheta,\omega) \coloneqq \frac{\alpha}{2}\|\vartheta\|_{L^2(0,T;\R^m)}^2+ \sum_{m\in[n_d]} g_m(\hat\bfx_{m,T}) + \beta\sum_{m\in[n_d]} \int_0^T \ell_m(\hat\bfx_{m,t}) \diff t.
\end{equation}
We define the expected objective over the batch realizations as
\begin{equation}\label{eq:objective_expected}
\overline J_h(\vartheta) \coloneqq \mathbb E_\omega[\hat J_h(\vartheta,\omega)].
\end{equation}
The random functional $\hat J_h$ corresponds to a single batch schedule (the fixed-schedule objective), whereas $\overline J_h$ is the expected objective associated with batch resampling.  The error bounds governing the optimization landscape are captured in the following theorem.

\begin{theorem}\label{thm:cost_consistency}
There exist constants $C_{\mathcal U}^{\rm str},C_{\mathcal U}^{\rm wk}>0$, independent of $h$, such that:
\begin{enumerate}
    \item Without additional assumptions,
    \begin{equation}\label{eq:strong_j_bound}
    \sup_{\vartheta\in\mathcal U} \mathbb E_\omega\left[ |\hat J_h(\vartheta,\omega)-J(\vartheta)|^2 \right] \le C_{\mathcal U}^{\rm str}h.
    \end{equation}
   \item If \Cref{ass:lipxdxF} holds, then
    \begin{equation}\label{eq:weak_j_bound}
    \sup_{\vartheta\in\mathcal U} |\overline J_h(\vartheta)-J(\vartheta)| \le C_{\mathcal U}^{\rm wk}h.
    \end{equation}
\end{enumerate}
\end{theorem}

The first estimate quantifies the batch-induced fluctuations of the randomized objective. It implies a RMS error of order $h^{1/2}$ and bounds the variance of the randomized objective. For the second, the expectation of the batching noise evaluated along the deterministic trajectory vanishes, while its correlation with the trajectory error leaves an $\mathcal{O}(h)$ contribution. Consequently, the expected objective satisfies an order-$h$ weak consistency estimate. This strong/weak rate separation is aligned with the classical convergence theory for RBMs; see \cite{JIN2020108877,MR4230431}.


Finally, we translate these consistency bounds to the optimal values and near-minimizers. Let us define
\begin{equation*}
V \coloneqq \inf_{\vartheta\in\mathcal U}J(\vartheta), \qquad \overline V_h \coloneqq \inf_{\vartheta\in\mathcal U}\overline J_h(\vartheta).
\end{equation*}

\begin{corollary}\label{cor:optimal_values}
Assume the hypotheses of \Cref{thm:cost_consistency}(ii) hold. Then $|\overline V_h - V| \le C_{\mathcal U}^{\mathrm{wk}}h$. Furthermore, the near-minimizers of both formulations satisfy:
\begin{itemize}
    \item Any $\varepsilon_h$-minimizer of $\overline J_h$ is a $(2C_{\mathcal U}^{\mathrm{wk}}h + \varepsilon_h)$-minimizer of $J$.
    \item Any $\varepsilon_h$-minimizer of $J$ is a $(2C_{\mathcal U}^{\mathrm{wk}}h + \varepsilon_h)$-minimizer of $\overline J_h$.
\end{itemize}
\end{corollary}

These estimates concern the statistical consistency of the optimization problem itself, rather than the convergence of any specific training algorithm. The tolerance $\varepsilon_h$ simply accounts for the residual error of the chosen numerical optimizer.

\begin{remark}[Fresh batch evaluations]\label{rem:new_realization}
In practice, a control trained on the expected objective is often evaluated
under a fresh batch schedule. Here, a fresh batch schedule means a newly
sampled sequence $\omega=(\omega_1,\ldots,\omega_{n_s})$, drawn from the
same sampling law and independently of all schedules used to train or
select the control. If $\vartheta_h \in \mathcal U$ is an $\varepsilon_h$-minimizer of $\overline J_h$, a direct application of the preceding bounds and the triangle inequality shows that its expected mean-square deviation from the true optimum under a fresh realization $\omega$ satisfies
\begin{equation*}
\mathbb E_\omega \left[ |\hat J_h(\vartheta_h, \omega) - V|^2 \right] \le 2C_{\mathcal U}^{\mathrm{str}}h + 2(2C_{\mathcal U}^{\mathrm{wk}}h + \varepsilon_h)^2.
\end{equation*}
\end{remark}

\begin{remark}[Ensemble averaging]\label{rem:ensemble}
If in training or evaluation one averages over $M$ independent batch schedules, defining $\hat J_{h,M}(\vartheta) \coloneqq \sum_{r=1}^M \hat J_h(\vartheta,\omega^{(r)})/M$, the standard bias-variance decomposition yields
\begin{equation*}
\sup_{\vartheta\in\mathcal U} \mathbb E\left[ |\hat J_{h,M}(\vartheta)-J(\vartheta)|^2 \right] \le \frac{C_{\mathcal U}^{\mathrm{str}}h}{M} + (C_{\mathcal U}^{\mathrm{wk}})^2 h^2,
\end{equation*}
where for the second term we assume the hypotheses of \Cref{thm:cost_consistency}(ii). Increasing $M$ reduces the variance term, while the limiting mean-square error as $M\to\infty$ is of order $\mathcal{O}(h^2)$.
\end{remark}

While the preceding results bound the error in the objective values, establishing convergence in the control space requires additional geometric properties of the objective landscape. A standard tool for this is the quadratic growth condition, which translates excess objective value into a distance to the solution set (see, e.g., \cite{DrusvyatskiyLewis2018}).

\begin{corollary}\label{cor:quadratic_growth}

Assume the hypotheses of \Cref{thm:cost_consistency}(ii) hold and that $\argmin_{\mathcal U}J \neq \varnothing$. Assume there exist constants $\delta,\gamma > 0$ such that for all $\vartheta \in \mathcal U$ with $J(\vartheta) \le V + \delta$,
\begin{equation*}
J(\vartheta) - V \ge \frac{\gamma}{2} \operatorname{dist}_{L^2}^2(\vartheta, \argmin_{\mathcal U}J).
\end{equation*}
If $h$ is sufficiently small so that $2C_{\mathcal U}^{\mathrm{wk}}h + \varepsilon_h \le \delta$, then any $\vartheta_h \in \mathcal U$ satisfying $\overline J_h(\vartheta_h) \le \overline V_h + \varepsilon_h$ obeys the bound
\begin{equation*}
\operatorname{dist}_{L^2}(\vartheta_h, \argmin_{\mathcal U}J) \le \sqrt{\frac{2}{\gamma}\left(2C_{\mathcal U}^{\mathrm{wk}}h + \varepsilon_h\right)}.
\end{equation*}

If, in addition, $J$ admits a unique minimizer
$\vartheta^\star\in\mathcal U$, then
\begin{equation}\label{eq:qg_control_unique}
\|\vartheta_h-\vartheta^\star\|_{L^2(0,T;\R^m)}
\le
\sqrt{
\frac{2}{\gamma}
\left(
2C_{\mathcal U}^{\mathrm{wk}}h+\varepsilon_h
\right)}.
\end{equation}

If furthermore \Cref{ass:lipthetaF} holds, then for every
$m\in[n_d]$ there exists a constant $C_{m,\mathcal U}>0$,
independent of $h$, such that
\begin{equation}\label{eq:trained_trajectory_consistency}
\max_{t\in[0,T]}
\E_\omega\left[
\|\bfx_{m,\vartheta^\star,t}
-\hat\bfx_{m,\vartheta_h,t}\|^2
\right]
\le
C_{m,\mathcal U}(h+\varepsilon_h).
\end{equation}

\end{corollary}



\section{Design}\label{sec:design}

The preceding analysis applies to any admissible random batch scheme. We now
investigate how the sampling strategy affects the approximation error and
the computational cost. We first compare the randomization constants
associated with different sampling schemes and then derive the resulting
cost--accuracy trade-off.

\subsection{Sampling scheme and error constant}\label{ss:varsamp}

Every admissible sampling scheme satisfies the order-$h$ mean-square bound in \Cref{th:convergence}. The sampling law enters the constant $\mathsf S$ in this bound, whose explicit expression is recalled below:
\begin{equation}\label{eq:S_factorized}
\mathsf S(\bfx_0,\vartheta)
=
2\sqrt{T}
\bigl(
\lambda_{\bfF,\bfx,\vartheta}\|\bfx_0\|
+\kappa_{\bfF,0,\vartheta}
\bigr)
\frac{e^{3\lambda_{\bfF,\bfx,\vartheta}T/\pi_{\min}}}{\pi_{\min}}
\|\Lambda(\bfx_0,\vartheta)\|_{L^1(0,T)}^{1/2}.
\end{equation}
This expression isolates the scheme dependence into two quantities: the minimal inclusion probability $\pi_{\min}$ and the variance $\Lambda_t$. A small $\pi_{\min}$ amplifies the exponential factor, while a large $\Lambda_t$ indicates that the randomized field fluctuates strongly around the full field. For a fixed initial datum and control, schemes with the same $\pi_{\min}$ can therefore be compared through their integrated field variance at the level of this bound. A smaller variance yields a smaller upper bound, but need not imply a smaller actual trajectory error: perturbations at different times and in different directions may be amplified differently by the dynamics.

To analyze the variance $\Lambda_t$, we define $\bff_{i,t} \coloneqq \bff_i(\bfx_t,\vartheta_{i,t})$. While the marginal probability $\pi_i$ controls individual neuron coverage, schemes with the same marginal inclusion probabilities may still yield substantially different variances due to the statistical dependence between neuron selections (the explicit general variance identity is deferred to \Cref{ss:proofs41}). 

To characterize these effects across different sampling strategies, we introduce the neuronal mean and variance along the trajectory:
\begin{equation*}
\bar\bff_t \coloneqq \frac{1}{p}\sum_{i\in[p]}\bff_{i,t}, \qquad \mu_t^2 \coloneqq \|\bar\bff_t\|^2, \qquad \sigma_t^2 \coloneqq \frac{1}{p}\sum_{i\in[p]}\|\bff_{i,t}-\bar\bff_t\|^2.
\end{equation*}
We classify sampling schemes into three main families. 
Their design quantities are summarized in \Cref{tab:as}.

\noindent\textbf{Family A: Uniform fixed-size sampling.} We sample uniformly among all subsets of a fixed size $r\in[p]$. Then $\pi_i=r/p$, and for $p \ge 2$ (the $p=1$ case is trivially deterministic), the variance evaluates to
\begin{equation*}
\Lambda_t = \frac{p^2(p-r)}{(p-1)r}\sigma_t^2.
\end{equation*}
This single family encompasses the deterministic full batch ($r=p$), the ``all-but-one'' scheme ($r=p-1$), and the ``pick-one'' scheme ($r=1$). Within this family, increasing the batch size $r$ monotonically increases $\pi_{\min}$ and decreases $\Lambda_t$, leading to a monotonic decrease in the constant in the trajectory-error bound.

\noindent\textbf{Family B: Fixed disjoint partition.} Assume that $r$ divides $p$, and let
$[p]=\mathcal B_1\sqcup\cdots\sqcup\mathcal B_{p/r}$ be a fixed partition into blocks of size $r$, and sample one block uniformly. Here $\pi_i = r/p$, but the variance heavily depends on the grouping. Defining the block mean $\bar\bff_{\mathcal{B}_j,t} \coloneqq \frac{1}{r}\sum_{i\in\mathcal{B}_j} \bff_{i,t}$, we have
\begin{equation}\label{eq:fixed_partition_variance}
\Lambda_t = p^2 \;\frac{r}{p} \sum_{j=1}^{p/r}
\|\bar\bff_{\mathcal{B}_j,t}-\bar\bff_t\|^2
\eqqcolon p^2 \sigma_{\mathcal{P},t}^2.
\end{equation}
For fixed disjoint batching, the error depends not only on the batch size but also on how neurons are grouped. A good partition is one for which every batch mean closely approximates the full neuronal mean. Thus, two schemes with the same $r$ and the same marginal probability $\pi_{\min}$ can yield substantially different errors depending on their dependence structure.

\noindent\textbf{Family C: Bernoulli dropout.} In standard dropout implementations, neurons are masked by i.i.d. random variables $b_i^{(k)} \sim \operatorname{Bernoulli}(q_B)$ with $q_B\in(0,1)$. This corresponds to sampling over all $2^p$ subsets of $[p]$ with specific probabilities.
\begin{proposition}\label{prop:bern-as-rbm}
Within the RBM framework \eqref{eq:Fomegaj}--\eqref{eq:hatFkt}, let the batches $\mathcal{B}_j$ range over all subsets of $[p]$ with sampling probability
\begin{equation}\label{eq:prob.bern}
\mathbb{P}(\omega=j) = q_B^{|\mathcal{B}_j|}(1-q_B)^{p-|\mathcal{B}_j|}.
\end{equation}
Then $\pi_i=q_B$ for every $i$, and \eqref{eq:RM} coincides with the standard Bernoulli dropout field $\hat{\bfF}_t = \frac{1}{q_B}\sum_{i\in[p]} b_i^{(k_t)}\bff_{i,t}$ for a.e. $t\in(0,T)$.
\end{proposition}
In this case, neuronal inclusions are entirely uncorrelated, yielding
\begin{equation*}
\Lambda_t = \frac{1-q_B}{q_B} p(\sigma_t^2+\mu_t^2).
\end{equation*}
Setting $q_B=1/2$ recovers uniform sampling over all possible subsets, while the limit $q_B \to 1$ naturally recovers the deterministic full-batch case.

\begin{table}[ht]
\centering
\renewcommand{\arraystretch}{1.3}
\begin{tabular}{lccl}
\toprule
\textbf{Scheme} & $\pi_{\min}$ & $\Lambda_t$ & \textbf{Interpretation} \\
\midrule
Full batch ($r=p$) & $1$ & $0$ & Deterministic limit \\
All-but-one ($r=p-1$) & $\frac{p-1}{p}$ & $\frac{p^2}{(p-1)^2}\sigma_t^2$ & Special case of uniform fixed-size \\
Pick-one ($r=1$) & $\frac{1}{p}$ & $p^2\sigma_t^2$ & Special case of uniform fixed-size \\
Uniform fixed-size ($r$) & $\frac{r}{p}$ & $\frac{p^2(p-r)}{(p-1)r}\sigma_t^2$ & Size $r$ sampled uniformly \\
Fixed disjoint partition & $\frac{r}{p}$ & $p^2 \sigma_{\mathcal{P},t}^2$ & Fixed partition into blocks of size $r$ \\
Bernoulli ($q_B$) & $q_B$ & $\frac{1-q_B}{q_B}p(\sigma_t^2+\mu_t^2)$ & All subsets ($q_B=1/2$ is uniform) \\
\bottomrule
\end{tabular}
\caption{Design quantities characterizing the error prefactor across sampling schemes.}
\label{tab:as}
\end{table}

To illustrate the impact of the correlation structure, consider comparing Bernoulli dropout against uniform fixed-size sampling when $q_B = r/p$. Both schemes have exactly the same marginal inclusion probability $\pi_i = r/p$ and thus the same stability factor in the bound. The difference lies entirely in the variance:
\begin{equation*}
\Lambda_t^{\rm Bernoulli} - \Lambda_t^{\textrm{Fixed size}} = \frac{p(p-r)}{r}\left(\mu_t^2 - \frac{\sigma_t^2}{p-1}\right).
\end{equation*}
Because Bernoulli sampling has random batch cardinality, its variance contains a contribution from the common mean component $\mu_t^2$. Fixed-size sampling explicitly removes this source of randomness. However, when the mean component is small relative to neuronal variability, Bernoulli dropout need not have a larger variance. In fact, uniform fixed-size sampling yields no larger integrated variance than Bernoulli dropout at the same marginal retention level if and only if $$\int_0^T \mu_t^2 \diff t \ge \frac{1}{p-1}\int_0^T \sigma_t^2 \diff t.$$

A similar comparison applies to a fixed disjoint partition. At the same
batch size $r$, its integrated variance is no larger than that of uniform
fixed-size sampling if and only if
\begin{equation*}
\int_0^T \sigma_{\mathcal P,t}^2\,\diff t
\leq
\frac{p-r}{(p-1)r}\int_0^T\sigma_t^2\,\diff t.
\end{equation*}
Thus, uniform fixed-size sampling provides a grouping-independent baseline.
Bernoulli sampling may yield a smaller integrated variance when the common mean component is small,
whereas a fixed partition has smaller variance only when its batches are
sufficiently balanced.

\subsection{Cost–accuracy trade-off}\label{sec:cost_accuracy}
While accuracy alone favors less aggressive randomization, computational efficiency depends on the trade-off between batch size and approximation error. This section analyzes this trade-off. We measure computational work by the proxy
\begin{equation*}
\mathsf W
\approx
(\text{number of numerical steps})
\times
(\text{expected number of neurons evaluated per step}).
\end{equation*}
This proxy relies on two standard simplifications: we ignore hardware-dependent effects and we assign unit cost to each neuron evaluation. 

Define the expected batch size by
\begin{equation*}
\bar r
\coloneqq
\E_\omega[|\mathcal B_\omega|]
=
\sum_{i=1}^p\pi_i.
\end{equation*}
For fixed-size schemes, $\bar r=r$, whereas for Bernoulli dropout $\bar r=pq_B$. For the remainder of this subsection, we fix the initial datum, control, and sampling scheme, and write $\mathsf S\coloneqq\mathsf S(\bfx_0,\vartheta)$ for the corresponding trajectory-error constant.

We align the numerical and switching grids by setting $\Delta t=\gamma h$, where $\gamma^{-1}\in\mathbb N$. Treating $h$ as a continuous design variable, the resulting work proxy for the random model is
\begin{equation}\label{eq:cost_proxy}
\mathsf W_{\rm R}(h)
\coloneqq
\frac{T\bar r}{\gamma h}.
\end{equation}
Let $\bfx_{t_k}^{\Delta t}$ and $\hat\bfx_{t_k}^{\Delta t}$ denote the numerical approximations of the full and randomized trajectories, respectively, obtained with time step $\Delta t$ and evaluated at the switching-grid points $t_k$. Since $\Delta t=\gamma h$ and $\gamma^{-1}\in\mathbb N$, every switching-grid point also belongs to the numerical grid.

To incorporate the actual numerical error, we assume that for the chosen ODE solver, both \eqref{eq:FM} and \eqref{eq:RM} admit first-order global error bounds:
\begin{equation}\label{eq:numerical_assumptions}
\left(\max_k\E_\omega\left[\|\hat\bfx_{t_k}-\hat\bfx_{t_k}^{\Delta t}\|^2\right]\right)^{1/2} \le c_{\mathrm{R}}\Delta t, \qquad \max_k\|\bfx_{t_k}-\bfx_{t_k}^{\Delta t}\| \le c_{\mathrm{F}}\Delta t,
\end{equation}
 where $c_{\mathrm{R}}, c_{\mathrm{F}} > 0$ are deterministic constants independent of $h$ and $\Delta t$. By \Cref{th:convergence} and the triangle inequality, the Root Mean Square (RMS) error between the exact full trajectory and the numerical random trajectory is bounded by
\begin{equation}\label{eq:ER_bound}
\mathcal{E}_{\mathrm{R}}(h) \coloneqq \underbrace{\vphantom{\sqrt{\mathsf{S} h}\gamma} \sqrt{\mathsf{S} h}}_{\text{randomization error}} + \underbrace{\vphantom{\sqrt{\mathsf{S} h}\gamma} c_{\mathrm{R}}\gamma h}_{\text{integration error}}.
\end{equation}
Within this work--accuracy model, we optimize the guaranteed work predicted by this upper bound. Accordingly, the optimal work and crossover tolerance derived below refer to this sufficient error constraint. They need not coincide with the minimal work or crossover obtained by comparing the methods at equal measured error. The continuous relaxation assumes the \emph{unsaturated regime}, where the optimal switching step satisfies $h^\star \le T$; otherwise, the true discrete work saturates at a single switching interval, corresponding to $1/\gamma$ numerical steps.

\begin{proposition}\label{prop:fixed_epsilon}
Assume $\mathsf S>0$ and $c_{\rm R}>0$. For a prescribed RMS tolerance $\varepsilon>0$, the largest admissible switching step, and hence the minimizer of $\mathsf W_{\rm R}(h)$ subject to $\mathcal E_{\rm R}(h)\le\varepsilon$, is
\begin{equation}\label{eq:optimal_h_rbm}
h^\star(\varepsilon)
=
\frac{4\varepsilon^2}{\mathsf S}
\left(
1+\sqrt{1+\frac{4c_{\rm R}\gamma\varepsilon}{\mathsf S}}
\right)^{-2}.
\end{equation}
The corresponding minimal work is
\begin{equation}\label{eq:optimal_cost_rbm}
\mathsf W_{\rm R}^\star(\varepsilon)
=
\frac{T\bar r\,\mathsf S}{4\gamma\varepsilon^2}
\left(
1+\sqrt{1+\frac{4c_{\rm R}\gamma\varepsilon}{\mathsf S}}
\right)^2.
\end{equation}
\end{proposition}

Defining the critical tolerance $\varepsilon_c \coloneqq \mathsf{S}/(c_{\mathrm{R}}\gamma)$, we can identify two asymptotic regimes corresponding to the dominant error mechanism:
\begin{enumerate}
\item \textbf{High accuracy (randomization-limited):} For $\varepsilon \ll \varepsilon_c$, the randomization error controls the permissible step size. We have $h^\star(\varepsilon) \sim \varepsilon^2/\mathsf{S}$ and $\mathsf W_{\mathrm{R}}^\star(\varepsilon) \sim T\bar r\mathsf{S}/(\gamma\varepsilon^2)$.
\item \textbf{Moderate tolerance (integration-limited):} For $\varepsilon \gg \varepsilon_c$, randomization is no longer the active constraint. We have $h^\star(\varepsilon) \sim \varepsilon/(c_{\mathrm{R}}\gamma)$ and $\mathsf W_{\mathrm{R}}^\star(\varepsilon) \sim T\bar r c_{\mathrm{R}}/\varepsilon$.
\end{enumerate}

To evaluate whether dropout is ultimately computationally advantageous, we compare it to the full model ($\mathsf{S}=0$). Respecting the saturation constraint $\Delta t_{\mathrm{F}} \le T$, the optimal time step is $\Delta t_{\mathrm{F}}^\star = \min(\varepsilon/c_{\mathrm{F}}, T)$, which yields the optimal work:
\begin{equation}\label{eq:optimal_cost_fm}
\mathsf W_{\mathrm{F}}^\star(\varepsilon) = p\max\left(\frac{T c_{\mathrm{F}}}{\varepsilon}, 1\right).
\end{equation}
To quantify this advantage, we define the \emph{asymptotic per-step computational gain}:
\begin{equation}\label{eq:definition_G}
G \coloneqq \frac{p}{\bar r} \frac{c_{\mathrm{F}}}{c_{\mathrm{R}}}.
\end{equation}
This constant weighs the raw reduction in neuron evaluations ($p/\bar r$) against the relative numerical stability of integrating the randomized dynamics ($c_{\mathrm{F}}/c_{\mathrm{R}}$).

\begin{proposition}\label{prop:relativecost}
Suppose $h_R^\star(\varepsilon) \le T$ and $\varepsilon/c_{\mathrm{F}} \le T$. Then, the optimal work ratio is given by
\begin{equation}\label{eq:ratio_exact}
\frac{\mathsf W_{\mathrm{F}}^\star(\varepsilon)}{\mathsf W_{\mathrm{R}}^\star(\varepsilon)} = G \frac{\sqrt{1+4\varepsilon/\varepsilon_c}-1}{\sqrt{1+4\varepsilon/\varepsilon_c}+1}.
\end{equation}
Consequently, within this regime:
\begin{itemize}
    \item If $G \le 1$, \eqref{eq:RM} is never computationally advantageous ($\mathsf W_{\mathrm{F}}^\star < \mathsf W_{\mathrm{R}}^\star$ for all valid $\varepsilon$).
    \item If $G > 1$, there exists a unique crossover tolerance 
    \begin{equation}\label{eq:crossover_tolerance}
    \varepsilon_\times \coloneqq \varepsilon_c \frac{G}{(G-1)^2},
    \end{equation}
    such that \eqref{eq:FM} is cheaper for $\varepsilon < \varepsilon_\times$, and \eqref{eq:RM} is cheaper for $\varepsilon > \varepsilon_\times$.
\end{itemize}
\end{proposition}

Equation \eqref{eq:ratio_exact} has two immediate consequences. As $\varepsilon\downarrow0$, the ratio $\mathsf W_{\mathrm F}^\star/\mathsf W_{\mathrm R}^\star$ tends to zero, so the \eqref{eq:FM} is asymptotically preferable. Conversely, when $\varepsilon/\varepsilon_c\gg1$, the ratio approaches $G$, which is the largest computational gain predicted by the model. Hence, within the unsaturated work--accuracy model, random batching is advantageous precisely when $G>1$ and $\varepsilon>\varepsilon_\times$.

The target accuracy also determines how candidates should be ranked. In the randomization-limited regime, the relevant quantity is $\bar r\mathsf S$; in the integration-limited one, it is $\bar r c_{\mathrm R}$. Thus, \Cref{ss:varsamp} compares batch compositions at a fixed size, while the present analysis determines how much randomization is worthwhile at the prescribed accuracy.

In summary, a computational advantage of \eqref{eq:RM} is possible at moderate tolerances only when the reduction in per-step work is sufficiently large ($G>1$) and the target tolerance exceeds $\varepsilon_\times$. Thus, an effective sampling design must jointly control batch size, inclusion probabilities, and field variance; reducing the batch size alone does not guarantee a net computational gain. These conclusions follow from the simplified computational-cost model considered here and should be validated for the numerical solver and hardware used in a concrete application.




\section{Numerical experiments}\label{sec:numerical_results}

In this section we validate the main results in this paper, namely, the trajectory convergence, sampling design, objective consistency and transport of measures, in a simple synthetic setup. We also compare the full and randomized models using the work proxy of \Cref{sec:cost_accuracy}. Implementation details are collected in \Cref{sec:numerical_details}.

\subsection{Trajectory convergence and sampling design}\label{subsec:num_trajectory}

The experiments in this subsection, as well as the objective-consistency and work-proxy evaluations that follow, share a common frozen classification model. We use \eqref{eq:node-p} with $T=1$, $d=2$, $p=24$, and $\sigma=\operatorname{GELU}$. The rows of $A_t\in\R^{p\times d}$ and columns of $W_t\in\R^{d\times p}$ collect $\mathbf a_{i,t}$ and $\mathbf w_{i,t}$, respectively. The control is parameterized by
\[
A_t=H_A(t;\eta_A),\qquad b_t=H_b(t;\eta_b),\qquad W_t=H_W(t;\eta_W),
\]
where each map is a neural network of one hidden layer with width $16$ and $\tanh$ activation. Regularization in \eqref{eq:objective_full} acts on the generated control $\vartheta^\eta$, rather than on the parameters $\eta$.

\begin{figure}[htbp]
\centering
\includegraphics[width=.32\linewidth]{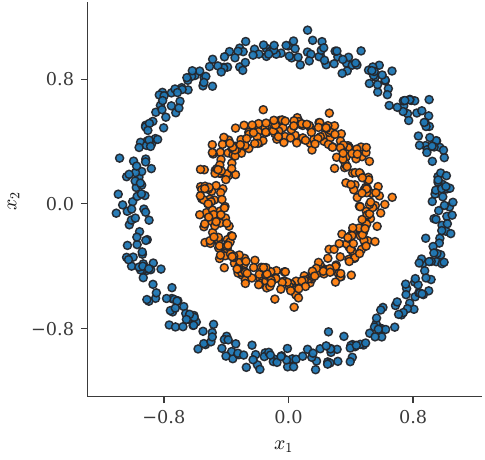}\hfill
\includegraphics[width=.32\linewidth]{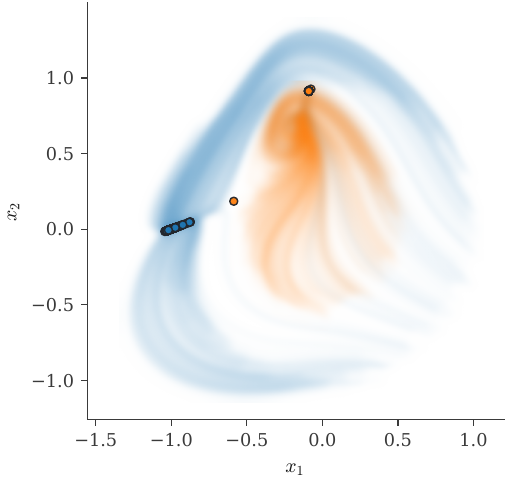}\hfill
\includegraphics[width=.32\linewidth]{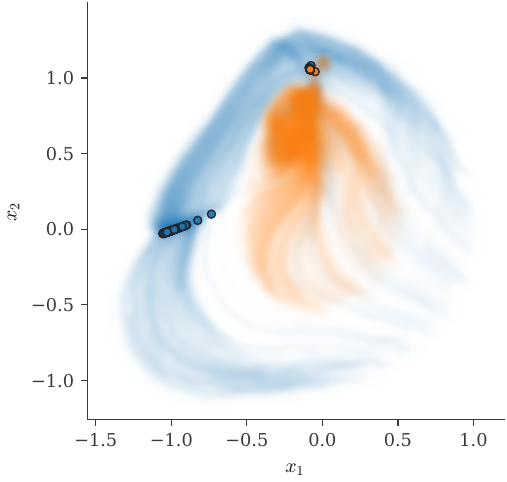}
\caption{Illustration of the classification flow. Left: the initial inputs (pooling the three splits). Centre and right: the full-model trajectories and one random-batch realization with uniform fixed-size sampling, $r=8$ and $h=2^{-11}$.  Blue and orange identify classes $-1$ and $+1$.}
\label{fig:classification_flow}
\end{figure}

We use independent training, calibration, and test splits from \texttt{make\_circles}. For labels $c_m\in\{-1,1\}$, the targets are $\bfy_m=(-1,0)$ when $c_m=-1$ and $\bfy_m=(0,1)$ otherwise, with $g_m(\bfx)=\ell_m(\bfx)=\|\bfx-\bfy_m\|^2$. The full model is trained once and then frozen. Trajectory, sampling-design, and work experiments use the test split; objective consistency uses the calibration split, with identical data in the full and randomized objectives. These experiments concern approximation at a fixed control, not convergence of a training algorithm or of minimizers. The resulting trajectories, for both the full model and a randomized realization, are illustrated in \Cref{fig:classification_flow}.

To evaluate trajectory convergence (as in  \Cref{th:convergence}), for test inputs $\bfx_0^m$ and independent schedules $\omega_k$, we compute
\begin{equation}\label{eq:empirical_trajectory_error}
\hat{\mathcal E}_{\mathrm{tr}}(h)
\coloneqq\frac{1}{n_{\mathrm{test}}}\sum_{m=1}^{n_{\mathrm{test}}}
\max_\ell\frac{1}{N_{\mathrm{sch}}}\sum_{k=1}^{N_{\mathrm{sch}}}
\|\bfx^m_{t_\ell}-\hat\bfx^{m,k}_{t_\ell}\|^2.
\end{equation}
The maximum is taken on a common time grid for every $h$. Refining this evaluation grid does not appreciably change the computed error. We compare uniform fixed-size sampling with $r=8$, Bernoulli sampling with $q_B=1/3$, and a contiguous partition into three blocks of size eight. All three have $\pi_i=1/3$ and expected batch size eight.

\begin{figure}[htbp]
\centering
\includegraphics[width=.85\linewidth]{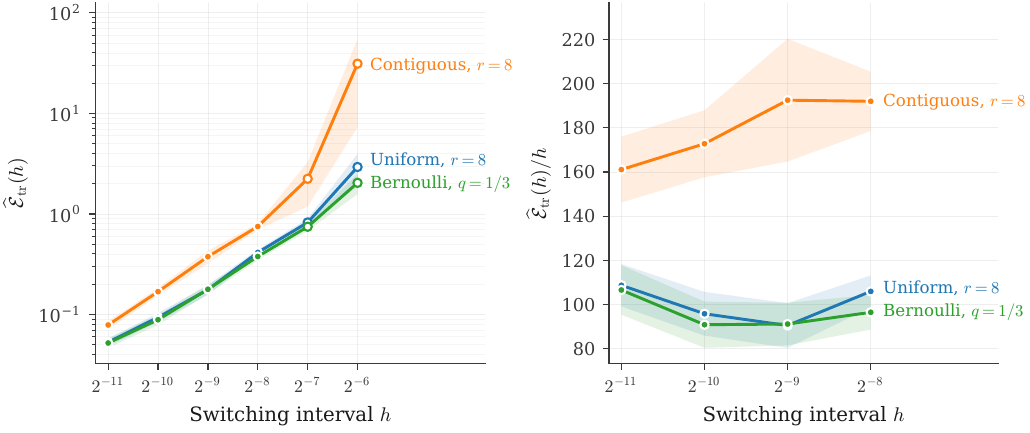}
\caption{Trajectory error \eqref{eq:empirical_trajectory_error} (left) and $\hat{\mathcal E}_{\mathrm{tr}}(h)/h$ over the fine range $h\le2^{-8}$ (right). Each scheme and switching scale uses $300$ schedules. Filled markers indicate the fine range; open markers are coarser diagnostic points. Bands are approximate $95\%$ grouped jackknife intervals over schedules. Least-squares slopes on the fine range are $0.98$ for uniform, $1.09$ for contiguous and $0.96$ for Bernoulli, with $R^2$ of $0.991$, $0.999$ and $0.994$.}
\label{fig:num_trajectory_convergence}
\end{figure}

In the fit range, \Cref{fig:num_trajectory_convergence} shows approximately order-$h$ decay and a comparatively stable rescaled error. Uniform and Bernoulli sampling give similar errors, whereas the contiguous partition produces errors about $1.5$--$2.1$ times larger. The coarser configurations depart from this behavior, particularly for the contiguous partition. These observations motivate a comparison with the field variance.

For a sampling law $D$, define the test-averaged integrated variance
\begin{equation}\label{eq:num_integrated_lambda}
\overline\Lambda_D\coloneqq\frac{1}{n_{\mathrm{test}}}
\sum_{m=1}^{n_{\mathrm{test}}}\|\Lambda^D(\bfx_0^m,\vartheta)\|_{L^1(0,T)}.
\end{equation}
\Cref{tab:variance_validation} compares the formulas in \Cref{tab:as} with Monte Carlo evaluations along the same full trajectories. The discrepancies are below $2\%$. The contiguous partition has nearly twice the integrated variance of the other two schemes.

\begin{table}[htbp]
\centering
\begin{tabular}{lrrr}
\toprule
Sampling scheme & Analytical $\overline\Lambda_D$ & Monte Carlo & Discrepancy (\%)\\
\midrule
Uniform fixed-size ($r=8$) &110.69&109.70&0.89\\
Bernoulli ($q_B=1/3$) &106.85&108.77&1.80\\
Contiguous partition ($r=8$) &208.27&209.46&0.57\\
\bottomrule
\end{tabular}
\caption{Integrated field variance, with $5000$ Monte Carlo draws per scheme. Inclusion probabilities and expected batch sizes are matched. Discrepancy is the absolute difference relative to the analytical value.}
\label{tab:variance_validation}
\end{table}

To examine the effect of grouping, we generate $100$ balanced partitions and evaluate both $\overline\Lambda_{\mathcal P}$ and $\hat{\mathcal E}^{\mathcal P}_{\mathrm{tr}}(h_{\mathrm{probe}})/h_{\mathrm{probe}}$ for every partition, with $h_{\mathrm{probe}}=2^{-8}$. Their integrated variances range from $19.9$ to $219.0$, with a strong empirical association with trajectory error (linear correlation $0.95$). The mean integrated variance is $109.2$, with standard error $5.8$, close to the uniform fixed-size value $110.7$. This agreement is expected: marginally, a block selected from a uniformly random balanced partition follows the uniform fixed-size distribution. The contiguous partition lies near the upper end of both distributions.

\begin{figure}[htbp]
\centering
\includegraphics[width=.4\linewidth]{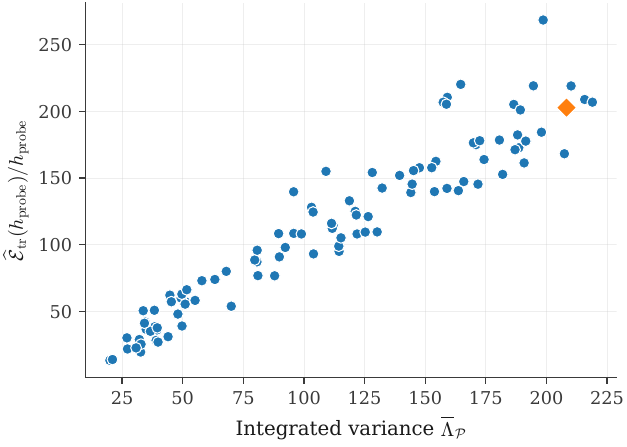}
\caption{Integrated variance and rescaled trajectory error for $100$ balanced partitions at $h_{\mathrm{probe}}=2^{-8}$, with $30$ schedules per partition. The diamond denotes the contiguous partition. }
\label{fig:partition_variance}
\end{figure}

These observations are consistent with the dependence of the error bound on the field variance. At matched inclusion probabilities, the stability factor in \eqref{eq:S_factorized} is common to all schemes, while the integrated field variance distinguishes their sampling designs. Larger integrated variance is strongly associated with larger trajectory error, as illustrated by \Cref{fig:partition_variance}.

\subsection{Objective consistency at a fixed control}\label{subsec:num_objective}

We normalize the objectives by the calibration sample size, writing $j=J/n_d$, $\hat j_h=\hat J_h/n_d$, and $\overline j_h=\overline J_h/n_d$. This changes only the constants in \Cref{thm:cost_consistency}. The control-energy term is evaluated identically and cancels in all differences. For the frozen control, we estimate the strong mean-square error and absolute weak bias by
\begin{align}
\hat{\mathcal E}_{J,\mathrm{str}}(h)
&\coloneqq\frac{1}{N_{\mathrm{sch}}}\sum_{k=1}^{N_{\mathrm{sch}}}|\hat j_h(\vartheta,\omega_k)-j(\vartheta)|^2,
\label{eq:empirical_strong_error}\\
\hat b_h&\coloneqq\frac{1}{N_{\mathrm{sch}}}\sum_{k=1}^{N_{\mathrm{sch}}}\hat j_h(\vartheta,\omega_k)-j(\vartheta),
\qquad\hat{\mathcal E}_{J,\mathrm{wk}}(h)\coloneqq|\hat b_h|.
\label{eq:empirical_weak_error}
\end{align}

Uncertainty in the weak error is assessed through the signed bias $\hat b_h$. The principal experiment uses uniform fixed-size sampling with $r=8$ and $768$ schedules per switching scale. We restrict the analysis to $h\le2^{-7}$: at $h=2^{-6}$, rare large objective values make the strong-error estimate highly uncertain, while the signed weak-bias interval contains zero.

\begin{figure}[htbp]
\centering
\includegraphics[width=\linewidth]{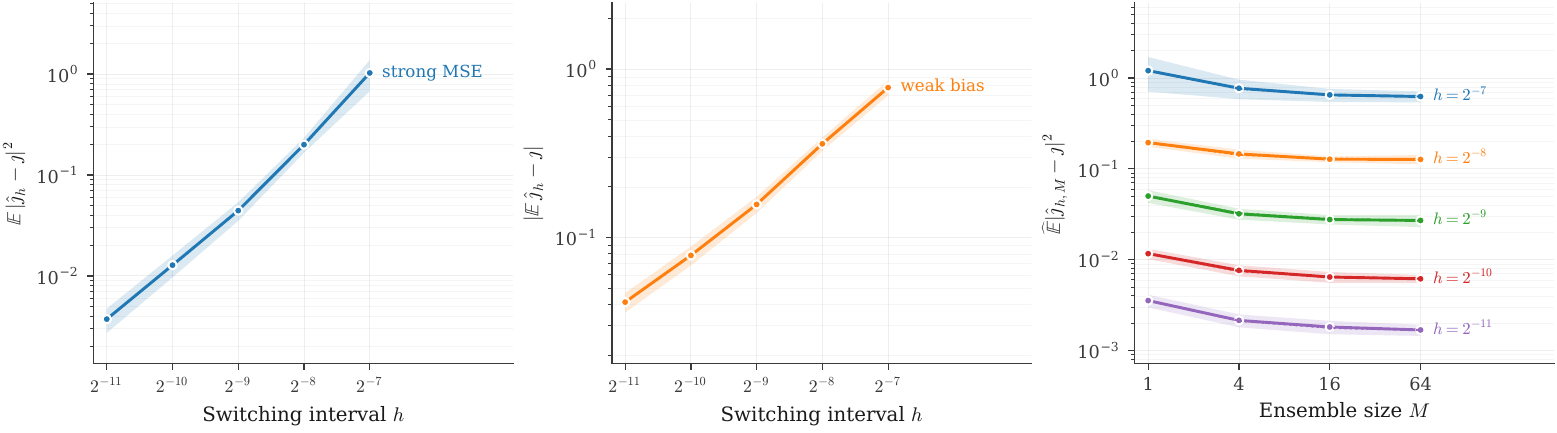}
\caption{Objective consistency at a fixed control, for uniform sampling with $r=8$ and $N_{\mathrm{sch}}=768$. Left: strong mean-square error, whose least-squares slope is $2.09$ with $R^2=0.995$. Centre: absolute weak bias, whose least-squares slope is $1.07$ with $R^2=0.999$. Right: ensemble mean-square error against $M$, labeled by $h$ at the end of each curve; the dotted curves are the empirical decomposition $\hat s_h^2/M+(\hat b_h^2-\hat s_h^2/N_{\mathrm{sch}})_+$ defined in the text. Shaded bands are approximate $95\%$ Monte Carlo intervals.}
\label{fig:objective_consistency}
\end{figure}

The weak bias decreases approximately linearly with $h$ in \Cref{fig:objective_consistency}, with a fitted slope of $1.07$, and its signed $95\%$ confidence interval excludes zero at every retained scale. The strong mean-square error decreases faster, with a fitted slope near two, while \Cref{thm:cost_consistency} guarantees an order-$h$ upper bound; the two estimates are therefore consistent with the theorem, which bounds the error rather than predicting its rate.

For averaging over $M$ schedules, we divide the available schedules into disjoint groups of size $M$ and compute the mean-square error of the corresponding group averages. The error satisfies the exact decomposition
\[
\E|\hat j_{h,M}-j|^2
=
\frac{\operatorname{Var}(\hat j_h)}{M}
+
(\overline j_h-j)^2
\le
C\left(\frac{h}{M}+h^2\right).
\]
Thus, averaging over more schedules reduces the variance by a factor $1/M$, while the bias remains unchanged. We compare the ensemble errors with $\hat s_h^2/M+(\hat b_h^2-\hat s_h^2/N_{\mathrm{sch}})_+$, where $\hat s_h^2$ is the sample variance of $\hat j_h$. The subtraction corrects the squared-bias estimate before truncation at zero. Groups are disjoint for each $M$, but different ensemble sizes share the same schedule pool. The curves illustrate variance reduction followed by saturation near the squared bias. As shown in \Cref{fig:objective_consistency}, when $h$ is sufficiently small, a moderate number of schedules already yields a small ensemble error.

The absolute size of the bias remains relevant. As shown in \Cref{tab:objective_relative_bias}, at the finest scale it is still about $29\%$ of the full objective, corresponding to a substantial relative perturbation.

\begin{table}[htbp]
\centering
\begin{tabular}{lrrrrr}
\toprule
$h$ &$2^{-7}$&$2^{-8}$&$2^{-9}$&$2^{-10}$&$2^{-11}$\\
\midrule
$|\hat b_h|/j$ &5.56&2.51&1.15&0.55&0.29\\
\bottomrule
\end{tabular}
\caption{Estimated relative weak bias for uniform fixed-size sampling. Entries are ratios, not percentages.}
\label{tab:objective_relative_bias}
\end{table}

\begin{figure}[htbp]
\centering
\includegraphics[width=.85\linewidth]{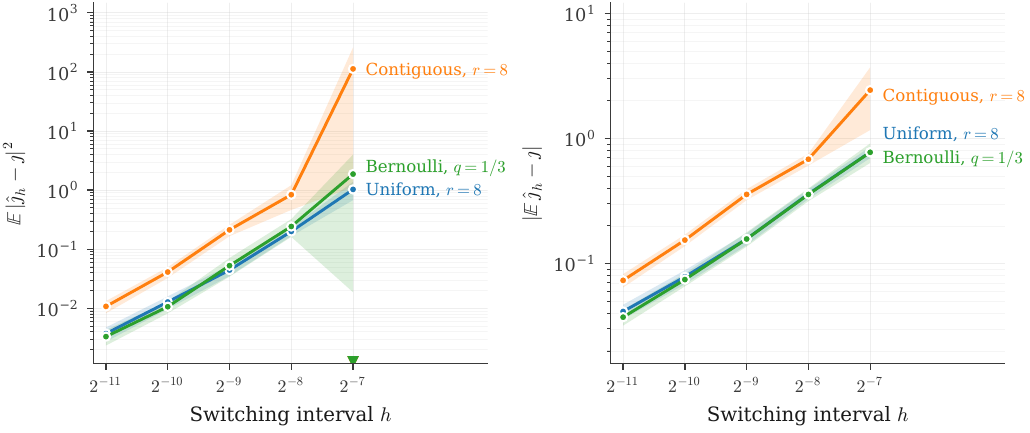}
\caption{Strong mean-square error (left) and absolute weak bias (right) for three sampling schemes, with $256$ schedules per scheme and switching scale. All schemes have $\pi_i=1/3$ and expected batch size eight. Curves are labeled at their ends and shaded bands are approximate $95\%$ Monte Carlo intervals; a downward triangle at the axis floor marks an estimate whose interval reaches zero.}
\label{fig:scheme_comparison}
\end{figure}

The comparison in \Cref{fig:scheme_comparison} shows similar fine-scale errors for uniform and Bernoulli sampling and larger errors for the contiguous partition. At coarse scales the schemes separate more strongly, particularly in the strong error. Thus, for this experiment, the sampling design has a visible effect on the objective error, with the contiguous partition consistently performing worse than the other two schemes.

\subsection{Transport of measures}\label{subsec:num_transport}

\begin{figure}[htbp]
\centering
\includegraphics[width=.4\linewidth]{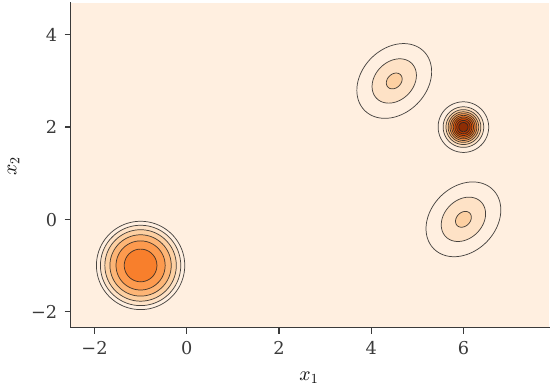}
\caption{The initial density $\rho_{\mathrm B}$ and the target mixture.}
\label{fig:transport_illustration0}
\end{figure}

We use a second neural ODE with $T=1$, $d=2$, $p=48$, $\sigma=\tanh$, and controls $A_t=A_0$, $W_t=W_0$, $b_t=b_0+b_1t$. The full field is trained by flow matching, \cite{lipman2023flow}, from the initial density
\[
\rho_{\mathrm B}(\bfx)=\frac{2}{\pi}\bigl(1-\|\bfx-(-1,-1)\|^2\bigr)_+,
\]
towards an equally weighted Gaussian mixture with means $(6,0)$, $(4.5,3)$ and $(6,2)$, whose third component is considerably narrower than the other two. The covariances, the flow-matching interpolant, and the remaining implementation details are given in \Cref{sec:numerical_details}, and both densities are represented in \Cref{fig:transport_illustration0}. The field is then frozen. Randomization uses uniform fixed-size sampling with $r=16$, hence $\pi_i=1/3$. We discretize the initial measure by $\sum_{m=1}^N\zeta_m\delta_{\bfx_0^m}$ using polar quadrature, with nonnegative weights summing to one. The full and randomized flows use the same initial points and weights.

\begin{figure}[htbp]
\centering
\includegraphics[width=.4\linewidth]{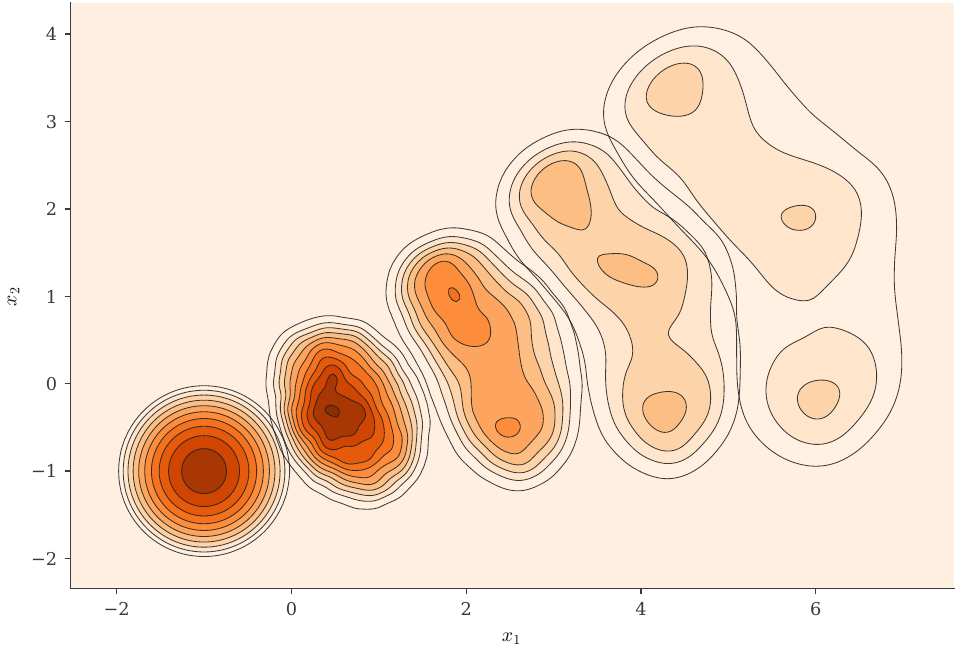}\hfill
\includegraphics[width=.4\linewidth]{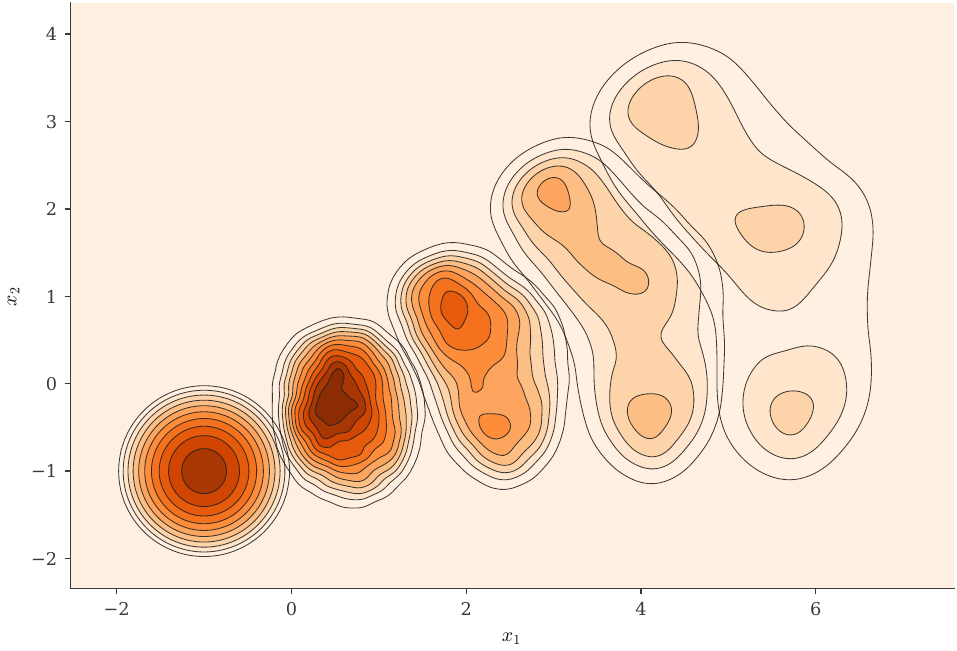}
\caption{The density transported by the full model (left) and by one random-batch realization (right) with $h=2^{-8}$ and $r=16$. Within each panel, five snapshots at $t=0, 0.25, 0.5, 0.75, 1$ are superimposed. The density at $t=0$ is $\rho_{\mathrm B}$ itself; the later snapshots are Gaussian KDEs computed from $20000$ trajectories. 
}
\label{fig:transport_illustration}
\end{figure}

As shown in \Cref{fig:transport_illustration}, both flows carry the initial ball along the same route and reach a comparable terminal spread. The learned field places the three modes of the target but does not fully concentrate the mass in them. This discrepancy is immaterial for the comparison below, which measures the randomized flow against the frozen full flow rather than against the target.

\paragraph{Coupling errors.}
We evaluate
\begin{align}
\hat{\mathcal E}_{\mathrm{sq}}(h)
&\coloneqq\frac{1}{N_{\mathrm{sch}}}\sum_{k=1}^{N_{\mathrm{sch}}}
\sum_{m=1}^N\zeta_m\|\Phi_T(\bfx_0^m)-\hat\Phi_T^{\omega_k}(\bfx_0^m)\|^2,
\label{eq:transport_sq_error}\\
\hat{\mathcal E}_{\mathrm{coup}}(h)
&\coloneqq\frac{1}{N_{\mathrm{sch}}}\sum_{k=1}^{N_{\mathrm{sch}}}
\left(\sum_{m=1}^N\zeta_m\|\Phi_T(\bfx_0^m)-\hat\Phi_T^{\omega_k}(\bfx_0^m)\|^2\right)^{1/2}.
\label{eq:transport_coup_error}
\end{align}
For each schedule, the square root defines the cost of the synchronous coupling between the discrete pushforward measures and therefore gives an upper bound on their $W_2$ distance. The corresponding bounds in \cref{th:convergence,rem:W2} have orders $h$ and $\sqrt h$, respectively.

\begin{figure}[htbp]
\centering
\includegraphics[width=.85\linewidth]{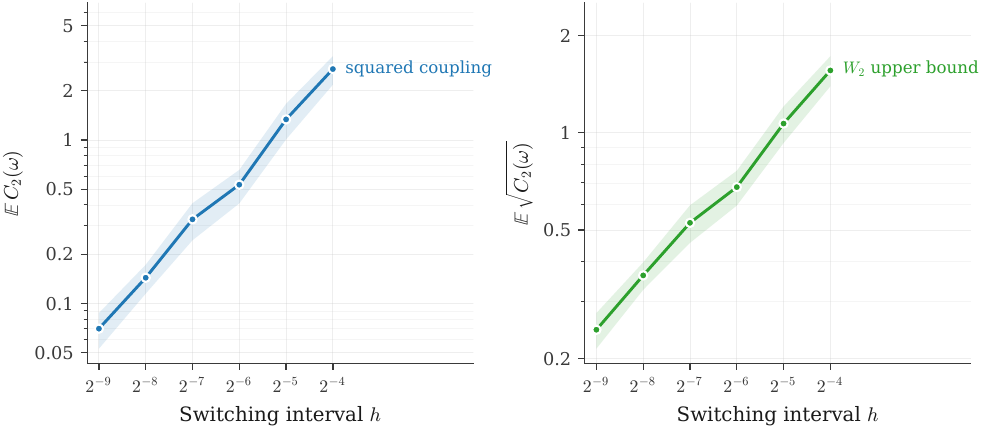}
\caption{Squared synchronous-coupling error \eqref{eq:transport_sq_error} (left) and mean synchronous-coupling upper bound \eqref{eq:transport_coup_error} (right). Bands are approximate $95\%$ intervals over schedules.}
\label{fig:transport_coupling}
\end{figure}

Both quantities decrease steadily under faster switching in \cref{fig:transport_coupling}. Least-squares log--log slopes over the tested range are $1.05$ for the squared coupling error and $0.53$ for the mean square-root coupling cost, close to the orders $h$ and $\sqrt h$ of \cref{th:convergence,rem:W2}. The right panel reports the $W_2$ upper bound induced by the synchronous coupling of particles starting from the same initial quadrature points.

\paragraph{Density errors.}
We evaluate densities by integrating the characteristics backward through each randomized schedule and using the divergence integral in \eqref{eq:Liouville}. The pointwise mean-square error is
\begin{equation}\label{eq:transport_pointwise_error}
\hat{\mathcal E}_{\rho,\mathrm{pw}}(h;\bfx^{(q)})
\coloneqq\frac{1}{N_{\mathrm{sch}}}\sum_{k=1}^{N_{\mathrm{sch}}}
|\rho_T(\bfx^{(q)})-\hat\rho_T^{\omega_k}(\bfx^{(q)})|^2.
\end{equation}
We choose five fixed points $y$ in the initial support with $\|y-c\|=0$, $0.4$, $0.6$, $0.8$, and $0.9$, where $c$ denotes the centre of the initial support, and evaluate the densities at their images under the full flow. These evaluation points are kept fixed across switching scales and schedules.

For the global density error, we estimate
\begin{equation}\label{eq:transport_L1_error}
\hat{\mathcal E}_{\rho,L^1}(h)
\coloneqq\frac{1}{N_{\mathrm{sch}}}\sum_{k=1}^{N_{\mathrm{sch}}}
\|\rho_T-\hat\rho_T^{\omega_k}\|_{L^1(\R^2)}.
\end{equation}
Since both $\rho_T$ and $\hat\rho_T$ have unit mass,
\begin{equation}\label{eq:l1_change_of_variables}
\|\rho_T-\hat\rho_T\|_{L^1(\R^d)}
=2\int_{\R^d}(\rho_T-\hat\rho_T)_+\,\diff\bfx
=2\E_{\mu_{\mathrm B}}\left[
\left(1-\frac{\hat\rho_T(\Phi_T(\bfy))}{\rho_T(\Phi_T(\bfy))}\right)_+
\right].
\end{equation}
The ratio is defined $\mu_{\mathrm B}$-almost everywhere. This identity accounts for mass outside the full-flow support while allowing integration on the initial quadrature cloud. The bracketed integrand lies in $[0,1]$, so the resulting estimate lies in $[0,2]$. Its nonsmoothness at density crossings motivates the spatial refinement check described below.

\begin{figure}[htbp]
\centering
\includegraphics[width=.96\linewidth]{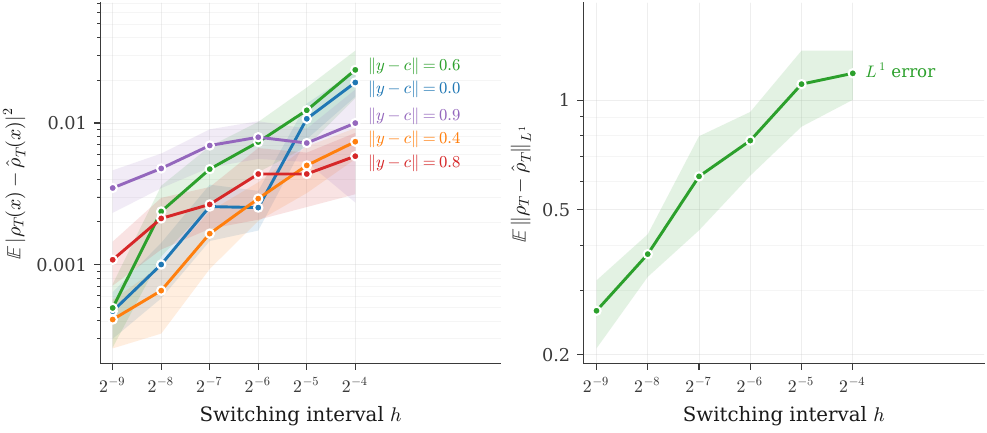}
\caption{Pointwise density mean-square errors at five fixed locations (left) and mean $L^1$ error (right). Labels indicate the distance of each point's preimage from the initial support centre. Bands are approximate $95\%$ intervals over schedules.}
\label{fig:transport_density}
\end{figure}

\Cref{fig:transport_density} shows that the pointwise convergence depends strongly on the evaluation location. At preimage radii $0$, $0.4$, and $0.6$, the fitted slopes are $1.06$, $0.87$, and $1.02$, respectively, close to the theoretical order $h$. At radii $0.8$ and $0.9$, closer to the boundary of the initial support, the decay is slower and less regular, with fitted slopes $0.46$ and $0.27$ and substantial overlap of the uncertainty bands. The measured decay is approximately linear at the three more interior locations, while the remaining points exhibit flatter finite-range trends. These observations do not establish the asymptotic pointwise rates.

The mean $L^1$ error decreases more regularly, with a fitted slope of $0.45$, close to the order-$\sqrt h$ reference. At the finest scale, the error remains of order a few tenths. Spatial and temporal refinement tests show very small sensitivity to the numerical discretization. Refining the spatial quadrature from $n=26$ to $n=37$ changes the estimated mean $L^1$ error by at most $0.12\%$ across the sweep. Halving the randomized characteristic step from $h/8$ to $h/16$ and the full reference step from $2^{-12}$ to $2^{-13}$ changes the estimate by at most $5.6\times10^{-8}$ in absolute value, with the reference-step refinement contributing at most $2.3\times10^{-15}$. These refinement discrepancies are therefore much smaller than the observed $L^1$ errors throughout the tested range.

\subsection{Work-proxy and accuracy}
\label{subsec:num_work_accuracy}

We compare computational saving and terminal trajectory accuracy using the
component-evaluation work proxy. For $p=24$, the randomized and full Euler
trajectories are evaluated on the same grid, with $\Delta t=h/4$, so that
the work saving is $1-r/p$. We also compare the minimum empirical work at
equal RMS accuracy for $p\in\{24,50,100,300\}$, resolving the full Euler
frontier with integer step counts in the competitive range.

\begin{figure}[htbp]
    \centering
    \begin{minipage}[c]{0.58\linewidth} 
        \centering
        \includegraphics[width=\linewidth]{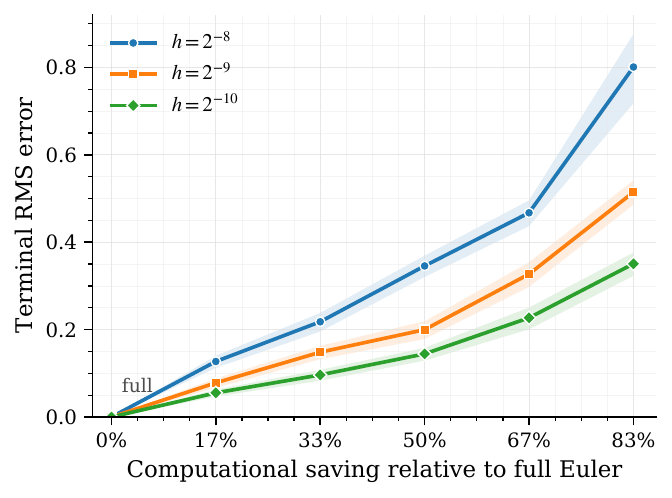}
    \end{minipage}%
    \hfill
    \begin{minipage}[c]{0.39\linewidth} 
        \centering
        \scriptsize
        \setlength{\tabcolsep}{3pt}
        \begin{tabular}{@{}c c c c c@{}} 
            \toprule
            $p$ & $(r,h)$ & $\varepsilon$
                & $(\mathsf W_{\mathrm R},\mathsf W_{\mathrm F})$
                & $\mathsf W_{\mathrm R}/\mathsf W_{\mathrm F}$ \\
            \midrule
            24 & $(20,2^{-5})$ & $0.457217$ & $(2560,1416)$ & $1.807910$ \\
            50 & $(40,2^{-3})$ & $0.476454$ & $(1280,650)$ & $1.969231$ \\
            100 & $(70,2^{-2})$ & $0.184291$ & $(1120,900)$ & $1.244444$ \\
            300 & $(150,2^{-2})$ & $0.165901$ & $(2400,2700)$ & $\mathbf{0.888889}$ \\
            \bottomrule
        \end{tabular}
    \end{minipage}

    \vspace{1em} 

    \begin{minipage}[t]{0.58\linewidth} 
        \captionof{figure}{
        Terminal RMS discrepancy versus the computational saving $1-r/p$.
        Curves correspond to different switching intervals $h$, and shaded
        regions are approximate $95\%$ intervals over $80$ batch schedules.
        }
        \label{fig:work_savings_precision}
    \end{minipage}%
    \hfill
    \begin{minipage}[t]{0.39\linewidth}
        \captionof{table}{
        Best empirical work ratios at equal terminal RMS accuracy.
        The pair in the fourth column reports randomized and full work,
        respectively.
        }
        \label{tab:equal_accuracy_refined}
    \end{minipage}
\end{figure}

\Cref{fig:work_savings_precision} shows the expected trade-off: increasing
the fraction of inactive components increases the terminal discrepancy,
while decreasing $h$ improves the approximation. At equal measured accuracy,
the configurations in \Cref{tab:equal_accuracy_refined} remain above the
unit work ratio for $p=24,50,100$. For $p=300$, the choice
$(r,h)=(150,2^{-2})$ reaches RMS tolerance $0.1659$ with
$\mathsf W_{\mathrm R}=2400$, compared with
$\mathsf W_{\mathrm F}=2700$, giving the observed ratio $0.889$.
This observed crossover is consistent with the cost--accuracy
mechanism described in \Cref{prop:relativecost}.

\section{Conclusions and discussion}\label{sec:conclusions}

We developed a random-batch framework for controlled differential systems
and quantified its approximation properties at the trajectory,
distribution, and optimization levels. On each fixed time interval
$[0,T]$, the randomized trajectories approximate the full dynamics with
mean-square error of order $\mathcal O(h)$. This yields Wasserstein
estimates and, under additional spatial regularity, pointwise density
estimates and global $L^1$ bounds. For training objectives, single-schedule
fluctuations are of order $\mathcal O(\sqrt h)$ in root mean square,
whereas the expected randomized objective has an $\mathcal O(h)$ weak
error. The latter estimate implies consistency of optimal values and
near-minimizers. Under quadratic growth and an optimization tolerance of
order $\mathcal O(h)$, the distance to the full minimizer set is of order
$\mathcal O(\sqrt h)$. When the full minimizer is unique, the additional
parameter-Lipschitz assumption yields mean-square trajectory consistency
at order $\mathcal O(h)$ for the corresponding controls.

We also compared sampling laws through their marginal inclusion
probabilities and correlations between component selections. The resulting
variance identities cover uniform fixed-size sampling, fixed partitions,
and Bernoulli dropout. They provide criteria for reducing the
trajectory-error bound, while the work--accuracy model determines the
switching scale that minimizes the work proxy subject to the prescribed
error bound. The numerical experiments examine trajectory and density errors, sampling
effects, and strong and weak objective consistency at a fixed control.
They are restricted to low-dimensional synthetic examples,
with the aim of isolating the mechanisms covered by the analysis rather
than assessing large-scale predictive performance.

\subsection*{Discussion and limitations}

\noindent\textbf{1. Existence of minimizers.}
The consistency estimates do not require attainment of the infima.
Existence is assumed whenever convergence to a minimizer is invoked.
For $\mathcal U=L^\infty(0,T;\mathsf K)$, compactness of
$\mathsf K$ alone does not ensure the compactness and continuity needed
for the direct method. Compact finite-dimensional control
parameterizations provide one setting in which existence can be obtained
from continuity. Another route is to impose temporal regularity through
a coercive $H^1$ penalty, using compactness in time and lower
semicontinuity; this changes the optimization problem being approximated.

\medbreak
\noindent\textbf{2. Fixed-schedule versus averaged training.}
Our optimization results concern $\overline J_h$. The estimate
$\sup_{\vartheta\in\mathcal U}\E_\omega
|\hat J_h(\vartheta,\omega)-J(\vartheta)|^2=\mathcal O(h)$
does not place the supremum inside the expectation and therefore does
not directly control a minimizer selected using the same schedules.
Fresh-schedule evaluation avoids this dependence for a given trained
control, but does not by itself establish that the control nearly
minimizes $\overline J_h$.

\medbreak
\noindent\textbf{3. Scope of the computational model.}
The work--accuracy analysis compares work proxies optimized under
sufficient error constraints. Its crossover tolerance concerns this
model and need not coincide with the crossover observed at equal measured
accuracy. Actual performance also depends on the integrator, sampling
overhead, control evaluation, and implementation. The first-order
integration-error assumption is additional to the continuous-time
convergence theory and must be checked for the controls and solver used.

\subsection*{Further directions}

A first direction is adaptive sampling based on the evolving trajectories.
Beyond minimizing the field variance, one could account
for how perturbations are amplified by the linearized flow, or weighted
by the adjoint associated with a training objective. Extending the
analysis would require conditional unbiasedness and control of the
resulting state-dependent fluctuations.

A second direction concerns training with finitely many schedules.
For compact finite-dimensional control parameterizations, covering
arguments combined with concentration estimates offer a possible route
to uniform objective bounds, with explicit dependence on the switching
scale, ensemble size, and parameter dimension. Such estimates could
connect the present consistency results to schedule-dependent
minimizers. Statistical generalization and subsequent model reduction
remain separate questions beyond the approximation theory developed here.

\appendix

\section{Proofs}\label{sec:proof_of_the_results}

To lighten the notation whenever the control $\vartheta$ is fixed or clear from context, we drop its explicit dependence and write $\bfF_t(\cdot)\coloneqq\bfF(\cdot,\vartheta_t)$ and $\hat{\bfF}_t(\cdot)\coloneqq\hat{\bfF}_t(\cdot,\vartheta_t)$. Under this convention, the spatial gradient is simply denoted by $\nabla$ instead of $\nabla_\bfx$. This shorthand applies to all analogous functions throughout the paper.

\subsection*{Preliminaries}\label{ss:prelim}
The following classic result establishes the existence and uniqueness of absolutely continuous solutions for ODEs that need not be continuous in time; see, e.g., \cite[Chapter~I]{hale_book_ode}.

\begin{lemma}[Carathéodory]\label{lemma:caratheodory}
Let $\mathbf{F}\colon[0,T]\times\R^d\to\R^d$ satisfy the following conditions:
\begin{enumerate}
    \item For almost every $t\in[0,T]$, the map $\bfx\mapsto\mathbf{F}_t(\bfx)$ is continuous.
    \item For each $\bfx\in\R^d$, the map $t\mapsto\mathbf{F}_t(\bfx)$ is measurable.
    \item There exists a function $m\in L^1(0,T)$ such that for almost every $t\in[0,T]$,
    \begin{equation}\label{eq:lingrowthcara}
    \|\mathbf{F}_t(\bfx)\|\leq m_t(\|\bfx\|+1) \qquad \text{for all } \bfx\in \R^d.
    \end{equation}
\end{enumerate}
Then, for any $\bfx_0\in\R^d$, there exists a solution $\bfx(\cdot)\in {AC}([0,T];\R^d)$ to
\begin{equation*}
 \begin{cases} 
    \dot\bfx_t=\mathbf{F}_t(\bfx_t), \qquad \text{for a.e. } t\in (0,T), \\
    \bfx_0\in \R^d.
 \end{cases}
\end{equation*}
Furthermore, if there exists a function $\lambda\in L^1(0,T)$ such that for almost every $t\in[0,T]$,
\begin{equation}\label{eq:uniqcara}
\|\mathbf{F}_t(\bfx)-\mathbf{F}_t(\bfy)\|\leq \lambda_t\|\bfx-\bfy\| \qquad \text{for all } \bfx,\bfy\in\R^d,
\end{equation}
then the solution is unique.
\end{lemma}

\begin{remark}\label{rema2}
Assume that \eqref{eq:uniqcara} holds. Then, the following statements are equivalent: 
\begin{enumerate}
    \item The linear growth bound \eqref{eq:lingrowthcara} is satisfied for some $m \in L^1(0,T)$.
    \item The map $t\mapsto\|\mathbf{F}_t(0)\|$ belongs to $L^1(0,T)$.  
\end{enumerate}
If either holds, \eqref{eq:lingrowthcara} is satisfied with $m_t=\lambda_t+\|\mathbf{F}_t(0)\|$.
\end{remark} 

To apply \Cref{lemma:caratheodory} to \eqref{eq:FM}, we set $\mathbf{F}_t(\bfx)\equiv\bfF(\bfx,\vartheta_t)$. The continuity of $\bfF$ and the measurability of $\vartheta$ naturally fulfill the Carathéodory conditions under the baseline regularity established in \Cref{ss:nodes}.

We recall Grönwall's inequality. For convenience, we state both the standard version and a useful modification for when the differential inequality naturally arises for $u^2$.

\begin{lemma}[Grönwall's inequality]\label{lemma:gronwall}
Let $t_0<t_1$, let $u\in{AC}([t_0,t_1];\R_{\geq0})$, and let $a,b\in L^1([t_0,t_1];\R_{\geq0})$. If either
\[
\frac{\diff}{\diff t}u_t\leq a_tu_t+b_t\qquad\text{or}\qquad\frac{\diff}{\diff t}u_t^2 \leq 2a_tu_t^2+2b_tu_t
\]
holds for almost every $t\in(t_0,t_1)$, then
\[
u_t \leq u_{t_0}e^{\int_{t_0}^t a_s\diff s} + \int_{t_0}^t b_s e^{\int_s^t a_r\diff r}\diff s
\]
for every $t\in[t_0,t_1]$.
\end{lemma}
\begin{proof}
The first implication is standard \cite[Appendix~B.2.j]{evans_pde}. To obtain the second, fix $\varepsilon>0$ and set $u_t^\varepsilon\coloneqq\sqrt{u_t^2+\varepsilon^2}$. Applying the first case to $u^\varepsilon$ and letting $\varepsilon\downarrow0$ yields the desired conclusion.
\end{proof}

We will use the following trajectory estimates.

\begin{lemma}\label{lem:estimation}
For any $\bfx_0\in \R^d$ and $\vartheta\in L^\infty(0,T;\Theta)$, the respective solutions $\bfx_t$ and $\hat{\bfx}_t$ of \eqref{eq:FM} and \eqref{eq:RM} satisfy
\begin{align}\label{eq:estimation_x}
\max_{t\in[0,T]}\|\bfx_t\|\leq \|\bfx_{0}\|e^{\lambda_{\bfF,\bfx,\vartheta} T}+\frac{\kappa_{\bfF,0,\vartheta}}{\lambda_{\bfF,\bfx,\vartheta}}\bigl(e^{\lambda_{\bfF,\bfx,\vartheta} T}-1\bigr)
\end{align}
(we adopt the convention that the fractional term equals $\kappa_{\bfF,0,\vartheta}T$ if $\lambda_{\bfF,\bfx,\vartheta}=0$), and, for every realization of the batch sequence,
\begin{align}\label{eq:alignpropest2}
\max_{t\in[0,T]}\|\bfx_t - \hat{\bfx}_t\| 
\leq e^{\frac{\lambda_{\bfF,\bfx,\vartheta}}{\pi_{\text{min}}} T}\Big(T\|\Lambda(\bfx_0,\vartheta)\|_{L^1(0,T)}\sum_{j\in[n_b]}q_j^{-1}\Big)^{1/2},
\end{align}
where $\lambda_{\bfF,\bfx,\vartheta}$ and $\kappa_{\bfF,0,\vartheta}$ are defined in \eqref{eq:condlipf} and \eqref{eq:condgrowthf}, $\pi_{\text{min}}=\min_{i\in[p]}\pi_i$ as in \eqref{eq:definition_pi}, $\Lambda$ as in \eqref{def:Lambda}, and $q_j$ in \eqref{def:qj}.

Moreover, assume \Cref{ass:lipthetaF} and let
$\vartheta_1,\vartheta_2\in L^\infty(0,T;\Theta)$. Let
$\mathsf B\subset\R^d$ and $\mathsf K\subset\Theta$ be compact sets containing both trajectories and the essential ranges of $\vartheta_{1}$ and $\vartheta_{2}$, respectively. Then
\begin{align}\label{eq:constant_lip_x}
\max_{t\in[0,T]}
\|\bfx_{\vartheta_1,t}-\bfx_{\vartheta_2,t}\|
\le
\lambda_{\bfF,\theta,\mathsf B,\mathsf K}
e^{(\lambda_{\bfF,\bfx,\vartheta_1}\wedge
\lambda_{\bfF,\bfx,\vartheta_2})T}
\|\vartheta_1-\vartheta_2\|_{L^1(0,T)}.
\end{align}
\end{lemma}

\begin{proof}[Proof of \Cref{lem:estimation}]
For a.e. $t\in(0,T)$, from \eqref{eq:FM} and using \eqref{eq:condgrowthf}, we get
\begin{align*}
    \frac{1}{2}\frac{\diff}{\diff t}\|\bfx_t\|^2 &= \langle \bfF_t(\bfx_t), \bfx_t \rangle\leq\|\bfF_t(\bfx_t)\|\cdot\|\bfx_t\|\leq \lambda_{\bfF,\bfx,\vartheta} \|\bfx_t\|^2 +\kappa_{\bfF,0,\vartheta} \|\bfx_t\|.
\end{align*}
By \Cref{lemma:gronwall}, for all $t\in[0,T]$
\begin{align*}
\|\bfx_t\| \leq \|\bfx_{0}\|e^{\lambda_{\bfF,\bfx,\vartheta} t}
+ \kappa_{\bfF,0,\vartheta}\int_0^t e^{\lambda_{\bfF,\bfx,\vartheta}(t-s)}\diff s \leq \|\bfx_{0}\|e^{\lambda_{\bfF,\bfx,\vartheta} T}
+ \frac{\kappa_{\bfF,0,\vartheta}}{\lambda_{\bfF,\bfx,\vartheta}}\bigl(e^{\lambda_{\bfF,\bfx,\vartheta} T}-1\bigr).
\end{align*}
To show \eqref{eq:alignpropest2}, for a.e. $t\in(0,T)$, let $k\equiv k_t$. From \eqref{eq:FM} and \eqref{eq:RM}, we get
\begin{align}\label{eq:alignpropest}
\nonumber\frac{1}{2}\frac{\diff}{\diff t}\|\bfx_t-\hat{\bfx}_t\|^2&=\langle \dot{\bfx}_t- \dot{\hat{\bfx}}_t,\bfx_t-\hat{\bfx}_t \rangle  = \langle \bfF_t(\bfx_t)-\bfF^{(\omega_k)}_t(\hat{\bfx}_t),\bfx_t-\hat{\bfx}_t  \rangle\\
\nonumber&= \langle \bfF_t(\bfx_t)-\bfF^{(\omega_k)}_t(\bfx_t),\bfx_t-\hat{\bfx}_t  \rangle+ \langle \bfF^{(\omega_k)}_t(\bfx_t)-\bfF^{(\omega_k)}_t(\hat{\bfx}_t),\bfx_t-\hat{\bfx}_t  \rangle\\
\nonumber&\overset{\text{(1)} }{\leq} \|\bfF_t(\bfx_t)-\bfF^{(\omega_k)}_t(\bfx_t)\|\cdot\|\bfx_t-\hat{\bfx}_t\|+\|\bfF^{(\omega_k)}_t(\bfx_t)-\bfF^{(\omega_k)}_t(\hat{\bfx}_t)\|\cdot\|\bfx_t-\hat{\bfx}_t\|\\
\nonumber&\leq \sum_{j\in[n_b]}\|\bfF_t(\bfx_t)-\bfF^{(j)}_t(\bfx_t)\|q_j^{1/2}q_j^{-1/2} \|\bfx_t-\hat{\bfx}_t\|+
\frac{\lambda_{\bfF,\bfx,\vartheta}}{\pi_{\text{min}}}\|\bfx_t-\hat{\bfx}_t\|^2\\
&\overset{\text{(2)} }{\leq}\Big(\Lambda_t(\bfx_0,\vartheta)\sum_{j\in[n_b]}q_j^{-1}\Big)^{1/2}\|\bfx_t-\hat{\bfx}_t\|+\frac{\lambda_{\bfF,\bfx,\vartheta}}{\pi_{\text{min}}}\|\bfx_t-\hat{\bfx}_t\|^2\nonumber\\
&=\beta_t\|\bfx_t-\hat{\bfx}_t\|+\frac{\lambda_{\bfF,\bfx,\vartheta}}{\pi_{\text{min}}}\|\bfx_t-\hat{\bfx}_t\|^2,
\end{align}
where we have used \eqref{eq:condlipf} and \eqref{eq:Fomegaj},
and applied Cauchy--Schwarz in \text{(1)} and \text{(2)}, and denoted
\begin{align*}
    \beta_t =  \Bigl(\Lambda_t(\bfx_0,\vartheta)\sum_{j\in[n_b]}q_j^{-1}\Bigr)^{1/2}.
\end{align*}
Since $\hat{\bfx}_0=\bfx_0$, applying \Cref{lemma:gronwall}
directly to \eqref{eq:alignpropest} yields, for every $t\in[0,T]$,
\begin{align*}
\|\bfx_t-\hat{\bfx}_t\|
&\le
\int_0^t
\beta_s
e^{\lambda_{\bfF,\bfx,\vartheta}(t-s)/\pi_{\min}}
\diff s \le
e^{\lambda_{\bfF,\bfx,\vartheta}T/\pi_{\min}}
\int_0^T\beta_s\,\diff s \\
&\le
e^{\lambda_{\bfF,\bfx,\vartheta}T/\pi_{\min}}
\Big(
T\|\Lambda(\bfx_0,\vartheta)\|_{L^1(0,T)}
\sum_{j\in[n_b]}q_j^{-1}
\Big)^{1/2},
\end{align*}
where the last inequality follows from Cauchy--Schwarz.

Finally, by \eqref{eq:estimation_x}, there exists a compact set
$\mathsf B\subset\R^d$ containing both trajectories
$\bfx_{\vartheta_1}$ and $\bfx_{\vartheta_2}$. Since $\vartheta_1,\vartheta_2\in L^\infty(0,T;\Theta)$, there exists
a compact set $\mathsf K\subset\Theta$ containing the essential ranges
of both controls. For a.e. $t\in(0,T)$,
\begin{align*}
\frac12\frac{\diff}{\diff t}
\|\bfx_{\vartheta_1,t}-\bfx_{\vartheta_2,t}\|^2
\le
\lambda_{\bfF,\bfx,\vartheta_1}
\|\bfx_{\vartheta_1,t}-\bfx_{\vartheta_2,t}\|^2 +
\lambda_{\bfF,\theta,\mathsf B,\mathsf K}
\|\vartheta_{1,t}-\vartheta_{2,t}\|
\|\bfx_{\vartheta_1,t}-\bfx_{\vartheta_2,t}\|.
\end{align*}
Hence, by \Cref{lemma:gronwall},
\[
\max_{t\in[0,T]}
\|\bfx_{\vartheta_1,t}-\bfx_{\vartheta_2,t}\|
\le
\lambda_{\bfF,\theta,\mathsf B,\mathsf K}
e^{\lambda_{\bfF,\bfx,\vartheta_1}T}
\|\vartheta_1-\vartheta_2\|_{L^1(0,T)}.
\]
Interchanging $\vartheta_1$ and $\vartheta_2$ gives the same estimate
with $\lambda_{\bfF,\bfx,\vartheta_2}$ in the exponential. Taking the
smaller of the two bounds proves \eqref{eq:constant_lip_x}.
\end{proof}

\subsection*{Proofs of \Cref{sec:convergence_dynamics}}\label{ss:prfs1}


\begin{proof}[Proof of \Cref{th:convergence}]
 To simplify notation, let $L\coloneqq\frac{\lambda_{\bfF,\bfx,\vartheta}}{\pi_{\min}}$ and $c\coloneqq\frac{\kappa_{\bfF,0,\vartheta}}{\pi_{\min}}.$ The growth bound gives $\|\hat{\bfF}_t(\bfx)\|\le L\|\bfx\|+c$ for a.e. $t$. Hence, by Grönwall's inequality (\Cref{lemma:gronwall}),
\begin{equation*}
\|\hat{\bfx}_t\| \le e^{Lt}\|\bfx_0\| + c\int_0^t e^{L(t-s)}\diff s,
\end{equation*}
which implies a uniform bound on the vector field along any random trajectory:
\begin{equation}\label{eq:thm1lastbound2}
\|\hat{\bfF}_t(\hat{\bfx}_t)\| \le e^{LT}\bigl(L\|\bfx_0\|+c\bigr) \eqqcolon M(\bfx_0,\vartheta), \qquad \text{for a.e. } t\in(0,T).
\end{equation}
(This formula remains valid if $L=0$). Now, for a.e. $t\in[0,T]$, let $k_t$ be the index such that $t\in[t_{k_t-1},t_{k_t})$, and define the residual vector field
\begin{equation*}
\delta_t \coloneqq \bfF_t(\bfx_t)-\bfF^{(\omega_{k_t})}_t(\bfx_t).
\end{equation*}
By adding and subtracting $\bfF^{(\omega_{k_t})}_t(\bfx_t)$, the derivative of the squared error is:
\begin{align*}
\frac12\frac{\diff}{\diff t}\|\bfx_t-\hat{\bfx}_t\|^2 &= \langle \bfF_t(\bfx_t)-\hat{\bfF}_t(\hat{\bfx}_t), \bfx_t-\hat{\bfx}_t \rangle \\
&= \langle \delta_t,\bfx_t-\hat{\bfx}_{t_{k_t-1}}\rangle + \langle\delta_t,\hat{\bfx}_{t_{k_t-1}}-\hat{\bfx}_t\rangle + \langle\bfF^{(\omega_{k_t})}_t(\bfx_t)-\hat{\bfF}_t(\hat{\bfx}_t), \bfx_t-\hat{\bfx}_t\rangle \\
&\le \langle\delta_t,\bfx_t-\hat{\bfx}_{t_{k_t-1}}\rangle + \|\delta_t\|\cdot\|\hat{\bfx}_t-\hat{\bfx}_{t_{k_t-1}}\| + L\|\bfx_t-\hat{\bfx}_t\|^2.
\end{align*}
Note that $\delta_t$ depends only on the current batch $\omega_{k_t}$, whereas the past state $\hat{\bfx}_{t_{k_t-1}}$ and the deterministic state $\bfx_t$ are independent of $\omega_{k_t}$. Since $\hat{\bfF}$ is unbiased, $\E_\omega[\delta_t]=0$, and thus the first term vanishes in expectation. Let
\begin{equation*}
u(t)\coloneqq \E_\omega\left[\|\bfx_t-\hat{\bfx}_t\|^2\right], \qquad r(t)\coloneqq \E_\omega\left[\|\hat{\bfx}_t-\hat{\bfx}_{t_{k_t-1}}\|^2\right].
\end{equation*}
Taking expectations, applying Cauchy-Schwarz to the second term (noting $\E_\omega[\|\delta_t\|^2] = \Lambda_t(\bfx_0,\vartheta)$), and multiplying by 2, we obtain the differential inequality:
\begin{equation}\label{eq:ddtexxrb}
u'(t) \le 2\Lambda_t(\bfx_0,\vartheta)^{1/2}r(t)^{1/2} + 2Lu(t), \qquad \text{for a.e. } t\in(0,T).
\end{equation}
Applying the Grönwall inequality (\Cref{lemma:gronwall}) yields
\begin{equation*}
u(t) \le 2\int_0^t e^{2L(t-s)}\Lambda_s(\bfx_0,\vartheta)^{1/2}r(s)^{1/2}\diff s,
\end{equation*}
and the Cauchy-Schwarz inequality to the integral over $[0,T]$ yields
\begin{equation}\label{eq:estimation_final}
u(t) \le 2e^{2LT}\|\Lambda(\bfx_0,\vartheta)\|_{L^1(0,T)}^{1/2}\left(\int_0^T r(s)\diff s\right)^{1/2}.
\end{equation}

To bound $r(s)$, let $s\in[t_{k-1},t_k)$. Using \eqref{eq:thm1lastbound2}, we have
\begin{equation*}
\|\hat{\bfx}_s-\hat{\bfx}_{t_{k-1}}\| = \left\|\int_{t_{k-1}}^s \hat{\bfF}_\tau(\hat{\bfx}_\tau)\diff\tau\right\| \le \int_{t_{k-1}}^s M(\bfx_0,\vartheta)\diff\tau \le M(\bfx_0,\vartheta)\, h.
\end{equation*}
Squaring and taking the expectation gives $r(s) \le M(\bfx_0,\vartheta)^2 h^2$. Integrating over $[0,T]$ yields
\begin{equation*}
\int_0^T r(s)\diff s \le T M(\bfx_0,\vartheta)^2 h^2.
\end{equation*}
Substituting this estimate into \eqref{eq:estimation_final}, we conclude that for all $t\in[0,T]$,
\begin{equation*}
\E_\omega\left[\|\bfx_t-\hat{\bfx}_t\|^2\right] \le \mathsf{S}(\bfx_0,\vartheta) \, h,
\end{equation*}
where the finite constant
\begin{equation}\label{eq:Cx0theta}
\mathsf{S}(\bfx_0,\vartheta) \coloneqq 2\sqrt{T}e^{2LT} M(\bfx_0,\vartheta) \|\Lambda(\bfx_0,\vartheta)\|_{L^1(0,T)}^{1/2}
\end{equation}
is explicitly independent of $h$. 

Finally, we note that $M(\bfx_0, \vartheta)$ and $\|\Lambda(\bfx_0, \vartheta)\|_{L^1(0,T)}^{1/2}$ grow at most linearly with $\|\bfx_0\|$ by virtue of \eqref{eq:thm1lastbound2}, \eqref{eq:condgrowthf}, and \eqref{eq:estimation_x}. Their product thus readily guarantees the bound \eqref{eq:S_quadratic}.
\end{proof}

\begin{proof}[Proof of \Cref{cor:convergence_different_control}]
By the uniform estimate in \Cref{th:convergence} and
\eqref{eq:constant_lip_x}, there exist constants $C_1,C_2>0$,
depending only on $\bfx_0$ and $\mathsf K$, such that
\[
\max_{t\in[0,T]}
\|\bfx_{\vartheta_1,t}-\bfx_{\vartheta_2,t}\|^2
\le
C_1\|\vartheta_1-\vartheta_2\|_{L^1(0,T)}^2
\]
and
\[
\max_{t\in[0,T]}
\E_\omega\left[
\|\bfx_{\vartheta_2,t}-\hat\bfx_{\vartheta_2,t}\|^2
\right]
\le
C_2 h.
\]
Using $\|a-b\|^2\le2\|a-c\|^2+2\|c-b\|^2$ and taking
$C_{\bfx_0,\mathsf K}\coloneqq2(C_1\vee C_2)$ proves the claim.
\end{proof}

\subsection*{Proofs of \Cref{ss:transp}}

\begin{proof}[Proof of \Cref{thm:transport}]
Fix $(t,\bfx)\in[0,T]\times\R^d$ and define the initial characteristics
\begin{equation*}
\bfy\coloneqq\Phi_t^{-1}(\bfx), \qquad \hat\bfy\coloneqq\hat\Phi_t^{-1}(\bfx).
\end{equation*}
To simplify notation, let $\lambda\coloneqq\lambda_{\bfF,\bfx,\vartheta}$ and $\hat\lambda\coloneqq\lambda_{\bfF,\bfx,\vartheta}/\pi_{\min}$. The inverse random flow $\hat\Phi_t^{-1}$ is almost surely Lipschitz continuous with constant $e^{\hat\lambda t}$. Thus, by adding and subtracting $\hat\Phi_t^{-1}(\Phi_t(\bfy))$, we can bound the backward error using the forward flow:
\begin{align*}
\|\bfy-\hat\bfy\| = \left\| \hat\Phi_t^{-1}(\hat\Phi_t(\bfy)) -\hat\Phi_t^{-1}(\Phi_t(\bfy)) \right\| \le e^{\hat\lambda t} \|\hat\Phi_t(\bfy)-\Phi_t(\bfy)\|.
\end{align*}
Applying \Cref{th:convergence} to the forward trajectories starting from $\bfy$, we deduce:
\begin{equation}\label{eq:bound_y_haty}
\E_\omega\left[\|\bfy-\hat\bfy\|^2\right] \le e^{2\hat\lambda T} \mathsf{S}(\bfy,\vartheta)\,h.
\end{equation}
Similarly, the forward flow $\hat\Phi_s$ is $e^{\hat\lambda s}$-Lipschitz for any $s\in[0,t]$. Applying the triangle inequality and \eqref{eq:bound_y_haty}, the distance between the characteristics at time $s$ satisfies:
\begin{align}\label{eq:bound_phis}
\E_\omega\left[\|\Phi_s(\bfy)-\hat\Phi_s(\hat\bfy)\|^2\right] &\le 2\E_\omega\left[\|\Phi_s(\bfy)-\hat\Phi_s(\bfy)\|^2\right] + 2e^{2\hat\lambda T} \E_\omega\left[\|\bfy-\hat\bfy\|^2\right] \nonumber\\
&\le 2\big(1+e^{4\hat\lambda T}\big) \mathsf{S}(\bfy,\vartheta)\,h.
\end{align}

Now, recall from \eqref{eq:Liouville} that we can write $\rho_t(\bfx)=\rho_{\rm B}(\bfy)e^A$ and $\hat\rho_t(\bfx)=\rho_{\rm B}(\hat\bfy)e^B$, where
\begin{equation*}
A \coloneqq -\int_0^t\nabla\cdot\bfF_s(\Phi_s(\bfy))\diff s, \qquad B \coloneqq -\int_0^t\nabla\cdot\hat{\bfF}_s(\hat\Phi_s(\hat\bfy))\diff s.
\end{equation*}
Notice that since $\rho_{\rm B}$ is globally Lipschitz and integrable, it is bounded, so $\rho_{\rm B}\in L^\infty(\R^d)$. Since $|A|\le d\lambda T$ and $|B|\le d\hat\lambda T$, the Lipschitz continuity of $\rho_{\rm B}$ and the mean value theorem applied to the exponential yield
\begin{equation}\label{eq:rho_diff_master}
|\rho_t(\bfx)-\hat\rho_t(\bfx)|^2 \le 2\lambda_{\rho_{\rm B}}^2 e^{2d\lambda T}\|\bfy-\hat\bfy\|^2 + 2\|\rho_{\rm B}\|_{L^\infty(\R^d)}^2 e^{2d\hat\lambda T}|A-B|^2.
\end{equation}

We decompose the exponent difference as $A-B=U+V$, where
\begin{align*}
U &\coloneqq -\int_0^t \nabla\cdot\left(\bfF_s(\Phi_s(\bfy)) -\hat{\bfF}_s(\Phi_s(\bfy))\right)\diff s, \\
V &\coloneqq -\int_0^t \nabla\cdot\left(\hat{\bfF}_s(\Phi_s(\bfy)) -\hat{\bfF}_s(\hat\Phi_s(\hat\bfy))\right)\diff s.
\end{align*}
For $V$, we use Cauchy-Schwarz and \Cref{ass:lipxdxF}. Combining this with \eqref{eq:bound_phis}, we obtain
\begin{equation}\label{eq:bound_V}
\E_\omega[|V|^2] \le \frac{d^2\lambda_{\nabla\bfF,\bfx,\mathsf K_\vartheta}^2 T}{\pi_{\min}^2} \int_0^t \E_\omega\left[\|\Phi_s(\bfy)-\hat\Phi_s(\hat\bfy)\|^2\right] \diff s \le C_1 \mathsf{S}(\bfy,\vartheta)\,h,
\end{equation}
where $C_1$ is independent of $h$, and the compact $\mathsf K_\vartheta\subset\Theta$ contains the essential range of $\vartheta$.

To estimate $U$, we split the integral. Let $I_k \coloneqq [t_{k-1},t_k)\cap[0,t]$ and define
\begin{equation*}
Z_k \coloneqq \int_{I_k} \nabla\cdot\left(\bfF_s(\Phi_s(\bfy)) -\bfF_s^{(\omega_k)}(\Phi_s(\bfy))\right)\diff s.
\end{equation*}
Because the expectation over the random batch is a finite sum,  $\E_\omega[\nabla\cdot\hat{\bfF}] = \nabla\cdot\bfF$. Since the deterministic path $\Phi_s(\bfy)$ is independent of $\omega_k$ and $\hat\bfF$ is unbiased, the random variables $Z_k$ are therefore independent and mean-zero. Furthermore, the divergence difference is bounded by $d\lambda(1+\pi_{\min}^{-1}) \eqqcolon D$, which implies $|Z_k| \le D|I_k|$. Therefore, the cross-terms in the squared sum vanish in expectation, yielding:
\begin{equation}\label{eq:bound_U}
\E_\omega[|U|^2] = \sum_k \E_\omega[|Z_k|^2] \le D^2\sum_k |I_k|^2 \le D^2 h \sum_k |I_k| \le D^2 T h.
\end{equation}

Using $|A-B|^2 = |U+V|^2 \le 2|U|^2+2|V|^2$, we inject \eqref{eq:bound_V} and \eqref{eq:bound_U} into \eqref{eq:rho_diff_master} alongside \eqref{eq:bound_y_haty} to conclude
\begin{equation}\label{eq:almost_done_rho}
\E_\omega\left[ |\rho_t(\bfx)-\hat\rho_t(\bfx)|^2 \right] \le C_2 \big(1+\mathsf{S}(\bfy,\vartheta)\big)h,
\end{equation}
where $C_2$ is independent of $h$. 

Finally, as established in \eqref{eq:S_quadratic}, $\mathsf{S}(\bfy,\vartheta)$ grows at most quadratically with its initial spatial argument: $\mathsf{S}(\bfy,\vartheta) \le C_3(1+\|\bfy\|^2)$. Since $\bfy=\Phi_t^{-1}(\bfx)$ and the inverse deterministic flow has at most linear growth, $\|\bfy\| \le C_4(1+\|\bfx\|)$. Substituting this into \eqref{eq:almost_done_rho} provides the final bound $\E_\omega[|\rho_t(\bfx)-\hat\rho_t(\bfx)|^2] \le C(\rho_{\rm B},\vartheta)\big(1+\|\bfx\|^2\big)h$.
\end{proof}

\begin{proof}[Proof of \Cref{cor:L1bound}]
The vector fields $\bfF$ and $\hat{\bfF}$ satisfy uniform linear growth bounds independent of the specific batch realization. Consequently, since the initial density $\rho_{\rm B}$ has a finite $q$-th moment, finite moments propagate forward in time. Thus, there exists $M_q>0$, independent of $h$ and the batch realization, such that
\begin{equation}\label{eq:MomentBound}
\sup_{t\in[0,T]} \int_{\R^d}\|\bfx\|^q\rho_t(\bfx)\diff\bfx + \sup_{t\in[0,T]} \int_{\R^d}\|\bfx\|^q\hat\rho_t(\bfx)\diff\bfx \le M_q.
\end{equation}
By Markov's inequality, this provides a uniform bound on the tail integrals that holds almost surely for any $R>0$:
\begin{equation}\label{eq:TailMoment}
\int_{\R^d\setminus B_R}\rho_t(\bfx)\diff\bfx + \int_{\R^d\setminus B_R}\hat\rho_t(\bfx)\diff\bfx \le \frac{M_q}{R^q} \quad \text{a.s.}
\end{equation}

For $B_R$, let $C_d$ denote the volume of the unit ball in $\R^d$, so $|B_R| = C_d R^d$. Using the Cauchy-Schwarz inequality in space and applying \Cref{thm:transport}, we obtain for any $R\ge1$:
\begin{align*}
\E_{\omega}\left[ \int_{B_R}|\rho_t(\bfx)-\hat\rho_t(\bfx)|\diff\bfx \right] 
&\le |B_R|^{1/2} \left( \int_{B_R} \E_{\omega}\left[ |\rho_t(\bfx)-\hat\rho_t(\bfx)|^2 \right]\diff\bfx \right)^{1/2} \\
&\le \sqrt{C_d R^d} \left( \int_{B_R} C(\rho_{\rm B},\vartheta)\big(1+\|\bfx\|^2\big)h \diff\bfx \right)^{1/2} \\
&\le \sqrt{C_d R^d} \left( C(\rho_{\rm B},\vartheta)\big(1+R^2\big) h \, C_d R^d \right)^{1/2}\le K \sqrt{h} \, R^{d+1},
\end{align*}
where $K>0$ is a constant independent of $h$, $R$, and $t$. Decomposing $\R^d = B_R \cup (\R^d\setminus B_R)$, taking the expectation over the batch realizations, and combining the core bound with \eqref{eq:TailMoment}, we get
\begin{equation*}
\E_{\omega}\left[\|\rho_t-\hat\rho_t\|_{L^1(\R^d)}\right] \le K\sqrt{h} \, R^{d+1} + \frac{M_q}{R^q}.
\end{equation*}
To balance these terms, we choose $R = 1\vee h^{-\frac{1}{2(q+d+1)}}$. Substituting this choice of $R$ into the bound and enlarging the constant if necessary finally yields:
\begin{equation*}
\E_{\omega}\left[\|\rho_t-\hat\rho_t\|_{L^1(\R^d)}\right] \le C(\rho_{\rm B},\vartheta,T,q) \, h^{\frac{q}{2q+2d+2}}.
\end{equation*}

If the initial density $\rho_{\rm B}$ is compactly supported, the linear growth
bounds for $\bfF$ and $\hat{\bfF}$ imply that there exists $R_\ast>0$,
independent of $h$, $t$, and the batch realization, such that
\[
\operatorname{supp}\rho_t\cup\operatorname{supp}\hat\rho_t
\subset B_{R_\ast}
\qquad\text{for all }t\in[0,T]
\]
almost surely. Applying the estimate above on $B_{R_\ast}$, while
the tail term vanishes, gives
\[
\E_\omega\left[\|\rho_t-\hat\rho_t\|_{L^1(\R^d)}\right]
\le K\sqrt h.
\]
\end{proof}

\subsection*{Proofs of \Cref{ss:train}}

\begin{proof}[Proof of \Cref{thm:cost_consistency}]
\textbf{Strong bound \eqref{eq:strong_j_bound}:} By the uniform version of \Cref{th:convergence}, for each $m\in[n_d]$
there exists $C_{m,\mathcal U}>0$ such that
\[
\sup_{\vartheta\in\mathcal U}
\max_{t\in[0,T]}
\E_\omega[
\|\hat\bfx_{m,\vartheta,t}-\bfx_{m,\vartheta,t}\|^2]
\le C_{m,\mathcal U}h.
\]

The control penalty cancels out when taking the difference of the objectives. Applying Minkowski's inequality in $L^2(\Omega)$, and using the fact that $\ell_m, g_m \in {C}^{1,1}_{\rm loc}(\R^d; \R_{\geq0})$ imply that $\ell_m$ and $g_m$ are locally Lipschitz for every $m\in[n_d]$, we have
\begin{align*}
\|\hat J_h(\vartheta,\cdot) - J(\vartheta)\|_{L^2(\Omega)} \le& \sum_{m\in[n_d]} \bigg[ \beta\lambda_{\ell_m,B_R} \int_0^T \|\hat\bfx_{m,t}-\bfx_{m,t}\|_{L^2(\Omega)} \diff t \\
& + \lambda_{g_m,B_R} \|\hat\bfx_{m,T}-\bfx_{m,T}\|_{L^2(\Omega)} \bigg],
\end{align*}
where $B_R \subset \R^d$ is a compact ball containing all trajectories uniformly. Substituting \eqref{eq:bound.th:convergence} into this expression gives $\|\hat J_h(\vartheta,\cdot) - J(\vartheta)\|_{L^2(\Omega)} \le \widetilde{C}_{\mathcal U}\sqrt{h}$. Squaring both sides yields \eqref{eq:strong_j_bound}.

\textbf{Weak bound \eqref{eq:weak_j_bound}:} Fix a deterministic control $\vartheta \in \mathcal U$ and let $e_{m,t} \coloneqq \hat\bfx_{m,t} - \bfx_{m,t}$. By Taylor expansion of the locally smooth losses,
\begin{align*}
g_m(\hat\bfx_{m,T}) - g_m(\bfx_{m,T}) &= \langle \nabla g_m(\bfx_{m,T}), e_{m,T} \rangle + \mathcal{O}(\|e_{m,T}\|^2), \\
\ell_m(\hat\bfx_{m,t}) - \ell_m(\bfx_{m,t}) &= \langle \nabla \ell_m(\bfx_{m,t}), e_{m,t} \rangle + \mathcal{O}(\|e_{m,t}\|^2).
\end{align*}
Since $\sup_{t} \mathbb E_\omega[\|e_{m,t}\|^2] \le C h$ and the trajectories remain in a compact ball $B_R$, the expected quadratic remainders are uniformly bounded by $\mathcal{O}(h)$. To control the expected linear terms, we introduce the deterministic adjoint state $\bfp_{m,t}$ associated with the full model, which solves
\begin{equation*}
\begin{cases}
\dot\bfp_{m,t} = -\nabla_\bfx\bfF(\bfx_{m,t},\vartheta_t)^\top\bfp_{m,t} - \beta\nabla\ell_m(\bfx_{m,t}), \hspace{1cm} \text{for a.e. } t\in (0,T),\\
\bfp_{m,T}=\nabla g_m(\bfx_{m,T}).
\end{cases}
\end{equation*}
Defining the constants $M_{g_m,R} \coloneqq \sup_{\bfx\in B_R} \|\nabla g_m(\bfx)\|$ and $M_{\ell_m,R} \coloneqq \sup_{\bfx\in B_R} \|\nabla \ell_m(\bfx)\|$, Grönwall's inequality guarantees that this adjoint state is uniformly bounded over $\mathcal{U}$:
\begin{equation*}
\sup_{\vartheta\in\mathcal U}\|\bfp_{m,\vartheta}\|_{L^\infty(0,T)} \le e^{\lambda_{\bfF,\bfx,\mathsf K}T}\left(M_{g_m,R} + \beta T M_{\ell_m,R}\right).
\end{equation*}
Since $e_{m,0}=0$, $\bfp_{m,T}=\nabla g_m(\bfx_{m,T})$, and $\beta\nabla\ell_m(\bfx_{m,t})
=
-\dot\bfp_{m,t}
-
\nabla_\bfx\bfF(\bfx_{m,t},\vartheta_t)^\top\bfp_{m,t},$ integration by parts gives
\begin{align*}
&\left\langle\nabla g_m(\bfx_{m,T}),e_{m,T}\right\rangle
+\beta\int_0^T
\left\langle\nabla\ell_m(\bfx_{m,t}),e_{m,t}\right\rangle\diff t
\int_0^T
\left\langle
\bfp_{m,t},
\dot e_{m,t}
-
\nabla_\bfx\bfF(\bfx_{m,t},\vartheta_t)e_{m,t}
\right\rangle\diff t.
\end{align*}
Summing over $m$ and taking expectations gives the total expected linear
contribution. To evaluate its integrand, we expand the randomized vector
field around the deterministic state:
\begin{align*}
\dot e_{m,t} - \nabla_\bfx\bfF(\bfx_{m,t},\vartheta_t)e_{m,t} &= \hat\bfF_t(\bfx_{m,t},\vartheta_t) - \bfF(\bfx_{m,t},\vartheta_t) \\
&\quad + \underbrace{\left( \nabla_\bfx\hat\bfF_t(\bfx_{m,t},\vartheta_t) - \nabla_\bfx\bfF(\bfx_{m,t},\vartheta_t) \right)}_{\coloneqq G_{m,t}}e_{m,t} + \mathcal{R}_{m,t}.
\end{align*}
By \Cref{ass:lipxdxF} on the compact parameter sets
$\mathsf K_i=\operatorname{proj}_i\mathsf K$, the Taylor remainder
satisfies uniformly
\[
\|\mathcal R_{m,t}\|
\le C_{\mathcal U}\|e_{m,t}\|^2.
\]
Thus, its expectation yields a $\mathcal{O}(h)$ contribution. 

The first term evaluates the difference of the fields at the deterministic state $\bfx_{m,t}$; by unbiasedness, its expectation is identically zero. To control the critical cross-term $G_{m,t}e_{m,t}$, we introduce the filtration $\mathcal F_k \coloneqq \sigma(\omega_1,\ldots,\omega_k)$. Here the control $\vartheta$ is deterministic and the current batch is independent of the past filtration. On any interval $t \in [t_{k-1}, t_k)$, we decompose the error as $e_{m,t} = e_{m,t_{k-1}} + (e_{m,t} - e_{m,t_{k-1}})$. Conditioning on the past yields
\begin{equation*}
\mathbb E_\omega\left[ G_{m,t} e_{m,t_{k-1}} \right] = \mathbb E_\omega\left[ \mathbb E_\omega\left[ G_{m,t} \mid \mathcal F_{k-1} \right] e_{m,t_{k-1}} \right] = 0,
\end{equation*}
because $\omega_k$ is independent of $\mathcal F_{k-1}$ and $\mathbb E_\omega[\nabla_\bfx\hat\bfF_t(\bfx,\theta)] = \nabla_\bfx\bfF(\bfx,\theta)$. The remaining part involves the increment $(e_{m,t} - e_{m,t_{k-1}})$, which satisfies
\begin{equation*}
\|e_{m,t} - e_{m,t_{k-1}}\| \le \|\hat\bfx_{m,t}-\hat\bfx_{m,t_{k-1}}\| + \|\bfx_{m,t}-\bfx_{m,t_{k-1}}\| \le \int_{t_{k-1}}^t \left(\|\hat\bfF_s\| + \|\bfF\| \right) \diff s \le C h,
\end{equation*}
uniformly bounded because the vector fields are uniformly bounded on $B_R \times \mathsf K$. Since both the adjoint state $\bfp_{m,t}$ and the gradient difference $G_{m,t}$ are uniformly bounded, the inner product with this increment contributes at most $\mathcal{O}(h)$ inside the integral. 

All constants above depend only on $\mathsf K$, the data, the losses, $T$, and the constants in \Cref{ass:lipxdxF}, and are therefore uniform over $\vartheta\in\mathcal U$. Summing over $m$ and taking the supremum over $\mathcal U$ proves \eqref{eq:weak_j_bound}.
\end{proof}

\begin{proof}[Proof of \Cref{cor:optimal_values}]
For any $\vartheta \in \mathcal U$, we have $\overline J_h(\vartheta) \ge J(\vartheta) - C_{\mathcal U}^{\mathrm{wk}}h \ge V - C_{\mathcal U}^{\mathrm{wk}}h$. Taking the infimum yields $\overline V_h \ge V - C_{\mathcal U}^{\mathrm{wk}}h$. The reverse inequality follows symmetrically, giving $|\overline V_h - V| \le C_{\mathcal U}^{\mathrm{wk}}h$. 

Let $\vartheta_h \in \mathcal U$ be an $\varepsilon_h$-minimizer of $\overline J_h$, meaning $\overline J_h(\vartheta_h) \le \overline V_h + \varepsilon_h$. We decompose the error evaluated on \eqref{eq:FM} as
\begin{equation*}
J(\vartheta_h) - V = \bigl(J(\vartheta_h) - \overline J_h(\vartheta_h)\bigr) + \bigl(\overline J_h(\vartheta_h) - \overline V_h\bigr) + \bigl(\overline V_h - V\bigr).
\end{equation*}
Applying the uniform bound \eqref{eq:weak_j_bound} to the first term, the near-minimizer assumption to the second, and the optimal value gap to the third gives $J(\vartheta_h) - V \le C_{\mathcal U}^{\mathrm{wk}}h + \varepsilon_h + C_{\mathcal U}^{\mathrm{wk}}h$. The symmetric statement for the minimizers of $J$ follows by exchanging the roles of $J$ and $\overline J_h$.
\end{proof}

\begin{proof}[Proof of \Cref{cor:quadratic_growth}]
By \Cref{cor:optimal_values}, $J(\vartheta_h)-V
\le
2C_{\mathcal U}^{\mathrm{wk}}h+\varepsilon_h.$ Since the right-hand side is at most $\delta$, the quadratic growth
condition applies, giving
\[
\operatorname{dist}_{L^2}^2
(\vartheta_h,\argmin_{\mathcal U}J)
\le
\frac{2}{\gamma}
\left(2C_{\mathcal U}^{\mathrm{wk}}h+\varepsilon_h\right).
\]

If $J$ admits the unique minimizer $\vartheta^\star$, then
$\operatorname{dist}_{L^2}
(\vartheta_h,\argmin_{\mathcal U}J)
=
\|\vartheta_h-\vartheta^\star\|_{L^2}$,
and \eqref{eq:qg_control_unique} follows. Assume furthermore \Cref{ass:lipthetaF}. Since
$\mathcal U=L^\infty(0,T;\mathsf K)$, applying
\Cref{cor:convergence_different_control} with
$\vartheta_1=\vartheta^\star$ and $\vartheta_2=\vartheta_h$ yields
\[
\max_{t\in[0,T]}
\E_\omega\left[
\|\bfx_{m,\vartheta^\star,t}
-\hat\bfx_{m,\vartheta_h,t}\|^2
\right]
\le
C_{\bfx_m,\mathsf K}
\left(
h+
\|\vartheta^\star-\vartheta_h\|_{L^1(0,T)}^2
\right).
\]
By Cauchy--Schwarz and \eqref{eq:qg_control_unique},
\[
\|\vartheta^\star-\vartheta_h\|_{L^1(0,T)}^2
\le
T\|\vartheta^\star-\vartheta_h\|_{L^2(0,T)}^2
\le
\frac{2T}{\gamma}
\left(2C_{\mathcal U}^{\mathrm{wk}}h+\varepsilon_h\right).
\]
Therefore,
\[
\max_{t\in[0,T]}
\E_\omega\left[
\|\bfx_{m,\vartheta^\star,t}
-\hat\bfx_{m,\vartheta_h,t}\|^2
\right]
\le
C_{m,\mathcal U}(h+\varepsilon_h),
\]
which proves \eqref{eq:trained_trajectory_consistency}.
\end{proof}
\subsection*{Proofs of \Cref{ss:varsamp}}\label{ss:proofs41}

\begin{proof}[Proof of \Cref{prop:bern-as-rbm}]
Let $S^{(k)}\coloneqq\{i\in[p]: b_i^{(k)}=1\}$. Since $b_i^{(k)}$ are i.i.d. Bernoulli$(q_B)$, for any subset $S\subseteq[p]$,
\begin{equation*}
\mathbb{P}\big(S^{(k)}=S\big) = \prod_{i\in S}\mathbb{P}(b_i^{(k)}=1) \prod_{i\notin S}\mathbb{P}(b_i^{(k)}=0) = q_B^{|S|}(1-q_B)^{p-|S|}.
\end{equation*}
Therefore, the inclusion probability is exactly the Bernoulli parameter: $\pi_i = \mathbb{P}(b_i^{(k)}=1) = q_B$. By definition of $k_t$, for a.e. $t\in(0,T)$ the randomized vector field is
\begin{equation*}
\hat{\bfF}_t = \sum_{i\in\mathcal{B}_{\omega_{k_t}}}\frac{1}{\pi_i}\bff_{i,t} = \frac{1}{q_B}\sum_{i\in[p]} b_i^{(k_t)}\bff_{i,t},
\end{equation*}
which coincides with the standard Bernoulli dropout field.
\end{proof}

\noindent\textbf{Derivation of the general variance identity.}
Let $I_i \coloneqq \mathbf{1}_{\{i\in\mathcal{B}_\omega\}}$ be the batch inclusion indicators, and $\pi_{ij} \coloneqq \mathbb{P}(i,j\in\mathcal{B}_\omega)$ the second-order inclusion probabilities (with $\pi_{ii}=\pi_i$). The randomized vector field is $\hat{\bfF}_t = \sum_{i\in[p]} \frac{I_i}{\pi_i}\bff_{i,t}$. Since $\E_\omega[I_i] = \pi_i$ and $\E_\omega[I_i I_j] = \pi_{ij}$, we expand the squared norm:
\begin{align*}
\Lambda_t &= \E_\omega\left[\left\|\sum_{i\in[p]} \left(\frac{I_i}{\pi_i}-1\right)\bff_{i,t}\right\|^2\right] = \sum_{i,j\in[p]} \E_\omega\left[\left(\frac{I_i}{\pi_i}-1\right)\left(\frac{I_j}{\pi_j}-1\right)\right] \langle\bff_{i,t},\bff_{j,t}\rangle \\
&= \sum_{i,j\in[p]} \frac{\E_\omega[I_i I_j] - \pi_i\pi_j}{\pi_i\pi_j} \langle\bff_{i,t},\bff_{j,t}\rangle = \sum_{i,j\in[p]} \frac{\pi_{ij}-\pi_i\pi_j}{\pi_i\pi_j} \langle\bff_{i,t},\bff_{j,t}\rangle.
\end{align*}

\noindent\textbf{Derivation for Uniform fixed-size sampling.}
Assuming $p \ge 2$, since $\mathcal{B}_\omega$ is chosen uniformly from all $\binom{p}{r}$ subsets of size $r$, we have $\pi_i = r/p$ and, for $i\neq j$, $\pi_{ij} = \frac{r(r-1)}{p(p-1)}$. Applying the general variance identity,
\begin{align*}
\Lambda_t &= \frac{1-r/p}{r/p}\sum_{i\in[p]}\|\bff_{i,t}\|^2 + \sum_{i\neq j} \frac{\frac{r(r-1)}{p(p-1)}-\frac{r^2}{p^2}}{r^2/p^2} \langle\bff_{i,t},\bff_{j,t}\rangle \\
&= \frac{p-r}{r}\sum_{i\in[p]}\|\bff_{i,t}\|^2 - \frac{p-r}{r(p-1)}\sum_{i\neq j} \langle\bff_{i,t},\bff_{j,t}\rangle.
\end{align*}
Substituting $\sum_{i\neq j} \langle\bff_{i,t},\bff_{j,t}\rangle = \|\sum_{i\in[p]} \bff_{i,t}\|^2 - \sum_{i\in[p]}\|\bff_{i,t}\|^2 = p^2\|\bar\bff_t\|^2 - \sum_{i\in[p]}\|\bff_{i,t}\|^2$, we obtain
\begin{align*}
\Lambda_t &= \frac{p-r}{r}\left(1+\frac{1}{p-1}\right)\sum_{i\in[p]}\|\bff_{i,t}\|^2 - \frac{p^2(p-r)}{r(p-1)}\|\bar\bff_t\|^2 \\
&= \frac{p(p-r)}{r(p-1)} \left( \sum_{i\in[p]}\|\bff_{i,t}\|^2 - p\|\bar\bff_t\|^2 \right) = \frac{p^2(p-r)}{(p-1)r} \sigma_t^2.
\end{align*}

\noindent\textbf{Derivation for Fixed disjoint partition.}
For a fixed partition of $p/r$ blocks where one block is sampled uniformly, $q_j = r/p$. The vector field is $\bfF^{(j)} = \frac{p}{r}\sum_{i\in\mathcal{B}_j} \bff_{i,t} = p\bar\bff_{\mathcal{B}_j,t}$. Thus,
\begin{equation*}
\Lambda_t = \sum_{j=1}^{p/r} q_j \|\bfF^{(j)} - \bfF_t\|^2 = \frac{r}{p} \sum_{j=1}^{p/r} \|p\bar\bff_{\mathcal{B}_j,t} - p\bar\bff_t\|^2 = p^2\sigma_{\mathcal{P},t}^2.
\end{equation*}

\noindent\textbf{Derivation for Bernoulli dropout.}
The variables $I_i$ and $I_j$ are independent for $i\neq j$, so $\pi_{ij} = \pi_i\pi_j$, eliminating the cross-terms in the variance identity. Since $\pi_i = q_B$, we immediately get
\begin{equation*}
\Lambda_t = \sum_{i\in[p]} \frac{1-q_B}{q_B}\|\bff_{i,t}\|^2 = \frac{1-q_B}{q_B}p(\sigma_t^2+\mu_t^2).
\end{equation*}

\subsection*{Proofs of \Cref{sec:cost_accuracy}}

\begin{proof}[Proof of \Cref{prop:fixed_epsilon}]
Because $\mathsf W_{\mathrm{R}}(h) = T\bar r/(\gamma h)$ is strictly decreasing with $h$ and $\mathcal{E}_{\mathrm{R}}(h) = \sqrt{\mathsf{S}h} + c_{\mathrm{R}}\gamma h$ is strictly increasing, the minimal work is attained precisely when $\mathcal{E}_{\mathrm{R}}(h^\star) = \varepsilon$. Let $z = \sqrt{h^\star}$. This yields the quadratic equation $c_{\mathrm{R}}\gamma z^2 + \sqrt{\mathsf{S}}z - \varepsilon = 0$. The unique positive root is
\begin{equation*}
z = \frac{\sqrt{\mathsf{S}+4c_{\mathrm{R}}\gamma\varepsilon}-\sqrt{\mathsf{S}}}{2c_{\mathrm{R}}\gamma} = \frac{\sqrt{\mathsf{S}}}{2c_{\mathrm{R}}\gamma}\left(\sqrt{1+\frac{4c_{\mathrm{R}}\gamma\varepsilon}{\mathsf{S}}}-1\right)=\frac{2\varepsilon}{\sqrt{\mathsf{S}}}\left(\sqrt{1+\frac{4c_{\mathrm{R}}\gamma\varepsilon}{\mathsf{S}}}+1\right)^{-1}.
\end{equation*}
Squaring this expression directly yields $h^\star(\varepsilon)$ as stated in \eqref{eq:optimal_h_rbm}. Substituting $h^\star(\varepsilon)$ into $\mathsf W_{\mathrm{R}}(h)$ provides \eqref{eq:optimal_cost_rbm}.
\end{proof}

\begin{proof}[Proof of \Cref{prop:relativecost}]
Dividing $\mathsf W_{\mathrm{F}}^\star(\varepsilon) = Tpc_{\mathrm{F}}/\varepsilon$ by \eqref{eq:optimal_cost_rbm}, we get
\begin{equation*}
\frac{\mathsf W_{\mathrm{F}}^\star(\varepsilon)}{\mathsf W_{\mathrm{R}}^\star(\varepsilon)} = \frac{4p c_{\mathrm{F}} \gamma \varepsilon}{\bar r \mathsf{S} \left(1+\sqrt{1+\frac{4c_{\mathrm{R}}\gamma\varepsilon}{\mathsf{S}}}\right)^2}.
\end{equation*}
Using $\varepsilon_c = \mathsf{S}/(c_{\mathrm{R}}\gamma)$ and the per-step gain $G = p c_{\mathrm{F}}/(\bar r c_{\mathrm{R}})$, this expression simplifies to
\begin{equation*}
\frac{\mathsf W_{\mathrm{F}}^\star(\varepsilon)}{\mathsf W_{\mathrm{R}}^\star(\varepsilon)} = G \frac{4\varepsilon/\varepsilon_c}{\left(1+\sqrt{1+4\varepsilon/\varepsilon_c}\right)^2}.
\end{equation*}
Applying the identity $\frac{a}{(1+\sqrt{1+a})^2} = \frac{\sqrt{1+a}-1}{\sqrt{1+a}+1}$ to $a = 4\varepsilon/\varepsilon_c$ establishes the exact ratio \eqref{eq:ratio_exact}.

Because the function $x \mapsto \frac{\sqrt{1+4x}-1}{\sqrt{1+4x}+1}$ is strictly monotonically increasing from $0$ to $1$ for $x \in (0,\infty)$, the ratio $\mathsf W_{\mathrm{F}}^\star/\mathsf W_{\mathrm{R}}^\star$ is strictly bounded above by $G$. Thus, if $G \le 1$, the ratio is strictly less than 1 for all finite $\varepsilon>0$. If $G > 1$, setting the ratio to 1 and solving for $\varepsilon/\varepsilon_c$ yields $\sqrt{1+4\varepsilon/\varepsilon_c} = \frac{G+1}{G-1}$. Squaring both sides and isolating $\varepsilon$ immediately provides the unique crossover tolerance $\varepsilon_\times = \varepsilon_c \frac{G}{(G-1)^2}$.
\end{proof}

\section{Implementation details}\label{sec:numerical_details}

Computations use double precision on a CPU. A root seed of $2026$ generates
separate seeds for data, initialization, schedules, partitions, and uncertainty
estimates. Unless stated otherwise, randomized trajectories use RK4 with
$\Delta t=h/8$. The complete code, configuration files, and additional
reproducibility information for all experiments are available at
\url{https://github.com/antonioalvarezl/2026-CTDropout-RBM/}.

\paragraph{Trajectory and sampling experiments.}

The training, calibration, and test splits contain $384$, $128$, and $256$ samples, with noise $0.05$ and inner-radius factor $0.5$. Training uses $800$ Adam steps at learning rate $2\times10^{-3}$, $\alpha=10^{-3}$, $\beta=10^{-1}$, and RK4 step $2^{-6}$. After each update, the representation parameters are projected onto $[-5,5]$. Running costs use the trapezoidal rule. Continuity of the parameterization on this bounded parameter set and $[0,T]$ yields a compact image; taking its closed convex hull gives a compact convex set $\mathsf K\subset\Theta$ containing all generated control values.

The common evaluation grid has spacing $2^{-9}$. Uncertainty uses a grouped
delete-one-group jackknife with $30$ groups of $10$ schedules. The slopes in
\cref{fig:num_trajectory_convergence} are least-squares log--log fits over
$h\le2^{-8}$. Full-reference trajectories use RK4 step $2^{-16}$; halving the
randomized integration step at the finest switching scale produces negligible
changes in the reported statistic. Integrated variances use a time grid of
step $2^{-8}$ and $5000$ Monte Carlo draws per scheme. Balanced partitions are
generated by uniform random permutations, using common block-selection
sequences across partitions.

\paragraph{Objective experiment.}
The final schedules are independent of a $256$-schedule pilot used for sample
sizing. The slopes in \cref{fig:objective_consistency} are least-squares
log--log fits over the range indicated there. Monte Carlo intervals use
$1.96$ standard errors; for the weak error, uncertainty is first computed for
the signed bias $\hat b_h$. The full objective reference uses RK4 step
$2^{-16}$, and the same reference grid is used for the control-energy term.

\paragraph{Transport experiment.}
The target-mixture covariances are
\[
\Sigma_1=\Sigma_2=
\begin{bmatrix}
0.2&0.05\\
0.05&0.2
\end{bmatrix},
\qquad
\Sigma_3=0.05I_2.
\]
We use the linear-interpolant flow-matching objective \cite{lipman2023flow}.
Independently sampling $\bfx_0\sim\rho_{\mathrm B}$, $\bfx_1$ from the target
mixture, and $t\sim\mathcal U(0,1)$, we set $\bfx_t=(1-t)\bfx_0+t\bfx_1$ and minimize
\[
\E\left\|
\bfF(\bfx_t,\vartheta_t)-(\bfx_1-\bfx_0)
\right\|^2.
\]
Training uses $12000$ Adam steps with minibatch size $1024$ and initial learning
rate $10^{-3}$, halved on plateau with patience $10$ and floor
$2\times10^{-5}$. The checkpoint with smallest validation mean-square error on
$4096$ independent samples is retained.

Polar quadrature uses $n$ Gauss--Legendre radial nodes and $4n$ equispaced
angles, with $n=16$ for the coupling quantities and $n=26$ for the global
density error. Backward characteristics use RK4 with $\Delta t=h/8$ and
full-reference step $2^{-12}$. Spatial refinement uses
$n\in\{13,26,37\}$ on common schedules. Intervals over schedules use
$1.96$ standard errors. Coupling and pointwise density errors use $40$ schedules per switching scale,
whereas the global $L^1$ error uses $12$.

\paragraph{Work and accuracy experiment.}
Full-model Euler steps range from $2^{-3}$ to $2^{-10}$, and the RK4 reference
uses step $2^{-16}$.




\bibliographystyle{abbrv} 
\bibliography{references.bib}

\end{document}